%% file: main.tex
\documentclass[10pt,twocolumn,letterpaper]{article}
\usepackage{multirow}
\usepackage{cvpr}              % To produce the CAMERA-READY version
\input{preamble}

\definecolor{cvprblue}{rgb}{0.21,0.49,0.74}
\usepackage[pagebackref,breaklinks,colorlinks,allcolors=cvprblue]{hyperref}

\def\paperID{2478} % *** Enter the Paper ID here
\def\confName{CVPR}
\def\confYear{2025}

\title{Patch Matters: Training-free Fine-grained Image Caption Enhancement via Local Perception}

\author{Ruotian Peng$^{1,4}$\footnotemark[1], 
Haiying He$^{2}$\footnotemark[1],
Yake Wei$^{3}$,
Yandong Wen$^{4}$,
Di Hu$^{3}$\footnotemark[2]\\
$^1$School of Future Technology, South China University of Technology\\
$^2$College of Science, China Agricultural University\\
$^3$Gaoling School of Artificial Intelligence, Renmin University of China\\
$^4$School of Engineering, Westlake University\\
{
\tt\small \{pengruotian, wenyandong\}@westlake.edu.cn}, 
{\tt\small 2963057784@cau.edu.cn},
{\tt\small \{yakewei, dihu\}@ruc.edu.cn} 
}
\begin{document}
\maketitle
\raggedbottom
\renewcommand{\thefootnote}{\fnsymbol{footnote}} %将脚注符号设置为fnsymbol类型，即特殊符号表示
\footnotetext[1]{Equal Contributions.} %对应脚注[1]
\footnotetext[2]{Corresponding authors.} %对应脚注[2]

\input{0_abstract}    
\input{1_intro}

\input{2_relatedwork}
\input{3_method}

\input{4_experiment}
\input{5_conclution}
\newpage
{
    \small
    \bibliographystyle{unsrt}
    \bibliography{main}
    % \bibliographystyle{unsrt}
    % \bibliography{bib/references}
}
\input{X_suppl}


\end{document}

%% file: preamble.tex
\usepackage{multirow}
\usepackage{tabularx}
\usepackage{colortbl}
\usepackage{stfloats}
\usepackage{tocloft} % 导入 tocloft 宏包，用于自定义目录样式

\usepackage{graphicx}

%% file: 0_abstract.tex
\begin{abstract}

High-quality image captions play a crucial role in improving the performance of cross-modal applications such as text-to-image generation, text-to-video generation, and text-image retrieval. To generate long-form, high-quality captions, many recent studies have employed multimodal large language models (MLLMs). However, current MLLMs often produce captions that lack fine-grained details or suffer from hallucinations, a challenge that persists in both open-source and closed-source models. Inspired by Feature-Integration theory, which suggests that attention must focus on specific regions to integrate visual information effectively, we propose a \textbf{divide-then-aggregate} strategy. Our method first divides the image into semantic and spatial patches to extract fine-grained details, enhancing the model's local perception of the image. These local details are then hierarchically aggregated to generate a comprehensive global description. To address hallucinations and inconsistencies in the generated captions, we apply a semantic-level filtering process during hierarchical aggregation. This training-free pipeline can be applied to both open-source models (LLaVA-1.5, LLaVA-1.6, Mini-Gemini) and closed-source models (Claude-3.5-Sonnet, GPT-4o, GLM-4V-Plus). Extensive experiments demonstrate that our method generates more detailed, reliable captions, advancing multimodal description generation without requiring model retraining. The source code are available at \url{https://github.com/GeWu-Lab/Patch-Matters}
% {$https://github.com/GeWuLab/Diagnosing_Relearning_ECCV2024$}
% %图像字幕一直以来在视觉理解任务上扮演者重要角色。如今的多模态大模型在大规模图像字幕数据上进行了训练，取得了不错的成果。但是目前的图像字幕数据集大多数由网上获取，面临着文本太短，质量低下和噪声多等问题，此外人工标注或者调用先进的闭源模型也是一笔巨大的资源开销。这一定程度限制了多模态大模型的发展。在这项工作中，我们提出了一个免于训练的生成图像字幕的方法。最大限度地激发了现有的多模态大语言模型的能力，从多个角度地提取视觉信息对应的文本描述。并且通过层级式的方法融合文本，在融合过程中为字幕增添细节和消除错误。大量的实验表明我们的方法不仅优于先前的字幕生成方法，能在多个闭源和开源模型上都能显著提高性能，且生成的字幕具有更长的长度和更多的细节信息。

% The ABSTRACT is to be in fully justified italicized text, at the top of the left-hand column, below the author and affiliation information.
% Use the word ``Abstract'' as the title, in 12-point Times, boldface type, centered relative to the column, initially capitalized.
% The abstract is to be in 10-point, single-spaced type.
% Leave two blank lines after the Abstract, then begin the main text.
% Look at previous \confName abstracts to get a feel for style and length.

% 图像字幕是衡量多模态大语言模型的理解能力的一个重要任务，同时还可以辅助生成高质量的图文数据集以供模型训练。但现有的模型和方法在生成细粒度和可靠的字幕上仍然存在困难。在这项工作中，我们通过免训练的方式最大化图像字幕的正确信息熵，去除错误信息，提供了更细粒度的字幕。具体来说，我们通过（1）空间和语义Patch切分（2）区域级确定性描述（3）分层融合的方式进行全局字幕的增强。显著增强了开源模型（llava1.5.llava1.6，Mini-Gemini）和闭源大模型（GPT4o,Claude 3.5, GLM -4v-plus）在详细字幕生成上的能力，Meteor，spice，bleu-4有7.11,3.96,3.78的增益

\end{abstract}

%% file: 1_intro.tex
\section{Introduction}
\label{sec:intro}

High-quality image descriptions can significantly enhance the performance of models across various domains, such as text-image retrieval \cite{zheng2025dreamlip,zhang2025long}, text-to-image \cite{DBLP:conf/iclr/ChenYGYXWK0LL24,zheng2024cogview3,betker2023improving}, and text-to-video generation \cite{polyak2024movie,yang2024cogvideox}. Research has shown that short and noisy descriptions often lead to erroneous outputs in models \cite{bai2024hallucination-survey} and hinder their ability to handle complex tasks \cite{bai2024survey}. In contrast, descriptions that includes rich visual elements and is free from hallucinations can effectively improve the model's ability to process complex scenarios \cite{zheng2025dreamlip,zhang2025long} and increase the reliability of its responses \cite{yu2024hallucidoctor}, ultimately boosting  model performance.
\begin{figure}
  \centering
   \includegraphics[width=1\linewidth]{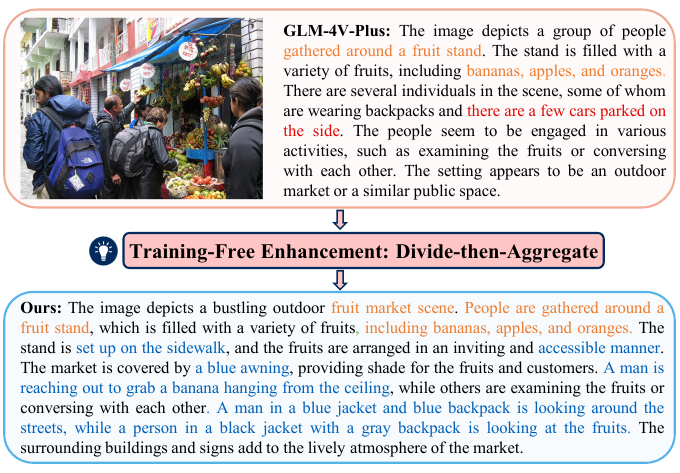}
    \caption{An illustration of current challenges faced by MLLMs in generating accurate image descriptions. Our approach provides a description with enhanced visual details and significantly reduced hallucinated content. \textcolor{orange}{Shared information}, \textcolor{cyan}{newly added details} and \textcolor{red}{hallucinations} are highlighted in different colors for clarity.}
    \label{fig:teaser1}
\end{figure}

Therefore, researchers have consistently sought ways to improve image captioning. Early methods \cite{vinyals2016show,you2016image} combined convolutional neural networks with recurrent neural networks for end-to-end training, enabling direct conversion from images to text. However, these methods had difficulty capturing the complex semantics of images \cite{anderson2018bottom} and often produced unnatural descriptions \cite{cheng2020stack}. The introduction of Transformer architectures \cite{DBLP:journals/tmlr/WangYHLLGLLW22,wang2022ofa} improved both image detail capture and the text coherence. More recently, multimodal large language models (MLLMs)—pre-trained on large image-text datasets and fine-tuned with specific instructions—have advanced image captioning further \cite{chen2023sharegpt4v,chen2024allava}, offering a new approach for automatically generating large-scale, long-form descriptions.

Despite these advancements, significant challenges persist in generating high-quality captions, particularly in terms of \textbf{\textit{lacking detail}} and \textbf{\textit{hallucination}}. Many models tend to focus on high-level scene descriptions, often overlooking fine-grained information such as actions or appearances, as shown in \cref{fig:teaser1}. On the other hand, when MLLMs aim to provide more detailed descriptions, they are more prone to hallucinations \cite{bai2024hallucination-survey, cui2023analysis-of-hallucination, wu2024relation-hallucination}, which result in captions that inaccurately reflect the image content. For example, some captions in \cref{fig:teaser1} describe non-existent objects or events. These challenges are present in both open-source and closed-source models, highlighting the need for new methods to effectively mitigate these issues, ideally in a training-free manner.

When observing images, the human brain extracts basic visual features in parallel, but to integrate these features into a coherent perception, attention must be sequentially directed to specific regions due to limited attentional capacity \cite{treisman1980feature}. Motivated by these insights, we propose a \textit{divide-then-aggregate} method that mimics this process. Our approach begins by dividing the image into patches to improve local perception, then aggregates this information to create a comprehensive global description. It is a training-free pipeline, applicable to both open-source and closed-source models. Specifically, we divide the image into spatial and semantic patches, preserving spatial relationships between nearby objects and enhancing key semantic features. These patches are used to generate multiple candidate captions, enriching local perception. To aggregate these details into a coherent global caption, we propose a two-stage hierarchical aggregation process. In the intra-patch aggregation stage, candidate descriptions within each patch are combined into a single, coherent caption. To address potential hallucinations and inconsistencies, we apply a semantic-level filtering strategy that classifies descriptions into three categories—\textit{same}, \textit{contradictory}, and \textit{unique}—based on semantic analysis. In the second stage of inter-patch aggregation, the semantic filtering strategy is applied again to resolve conflicts across patch-level descriptions, ultimately helping to form a detailed global caption.

We conduct extensive experiments to evaluate our pipeline on both open-source models such as LLaVA-1.5 \cite{liu2024improved}, LLaVA-1.6 \cite{liu2024improved}, and Mini-Gemini \cite{li2024mini-gemini}, as well as closed-source models like Claude-3.5-sonnet \cite{claude3.5}, GPT-4o \cite{hurst2024gpto}, and GLM-4V-Plus \cite{glm2024chatglm} across multiple benchmark datasets. Our results consistently demonstrate that our method generates more detailed and reliable captions, \eg, +4.7 in CIDEr for LLaVA and +2.07 in CIDEr for GPT-4o. In summary, our main contributions are as follows:

\begin{itemize}
    % \item We developed a training-free pipeline that enhances image captioning for both open-source and closed-source models by proposing a \textit{divide-then-aggregate} approach, which captures fine-grained details from spatial and semantic patches and hierarchically aggregates them with semantic filtering to mitigate hallucinations and improve global descriptions.
     \item  We propose a \textit{\textbf{divide-then-aggregate}} approach that captures details from spatial and semantic patches, and hierarchically aggregates them with semantic filtering to enhance the overall quality of the captions.
    \item The developed method is a training-free pipeline that enhances image captioning, applicable to both open-source and closed-source models.
    \item We conducted experiments on multiple models across several benchmark datasets, consistently demonstrating that our method increases fine-grained details and mitigates hallucinations in the captions.
\end{itemize}

\begin{figure*}
  \centering
  \setlength{\belowcaptionskip}{-5mm}
  \vspace{0mm}
   \includegraphics[width=0.92\linewidth]{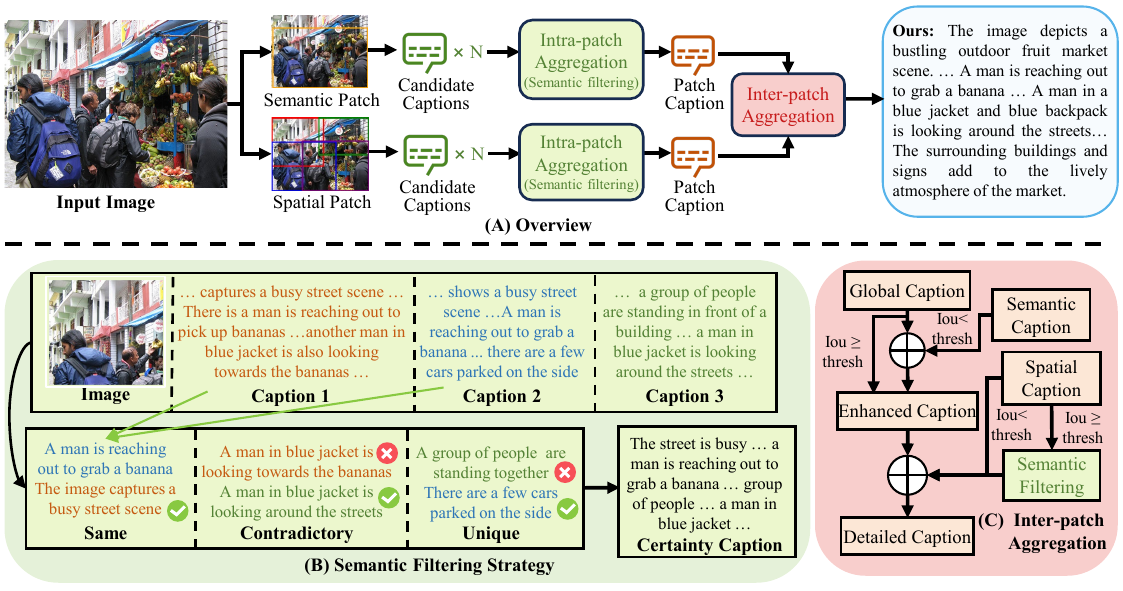}

    \caption{Our train-free \textbf{divide-then-aggregate} pipeline. (A) \textbf{Overview}: We divide the image into spatial and semantic patches to capture finer details (\cref{Sec3.2:Patch Slicing}), and perform hierarchical aggregation (intra-patch and inter-patch) to generate detailed and reliable image captions (\cref{Sec3.3:Hierarchical aggregation}). (B) \textbf{Semantic Filtering}: During aggregation, candidate descriptions are classified into same, contradictory, and unique categories, which are then consolidated into a coherent caption to mitigate hallucination and prevent information conflict. (C) \textbf{Inter-patch Aggregation}: When fusing descriptions from different patches, we use IoU to determine assess whether semantic patch enhancement is required to incorporate global information, and whether Semantic Filtering is needed to prevent conflicts.}
    \label{fig:method}
\end{figure*}

%% file: 2_relatedwork.tex
\section{Related Work}
\label{sec:formatting}
% \subsection{Image Captioning}
\subsection{Multi-Modal Large Language Models}

In recent years, large language models (LLMs) \cite{dubey2024llama,brown2020language,vicuna2023} have made significant advancements in the field of natural language processing, sparking efforts to extend their capabilities to multimodal tasks. This development has led to the emergence of MLLMs \cite{liu2024llava,hurst2024gpto,li2024mini-gemini,du2025crabunifiedaudiovisualscene,wei2022learning} , which can process visual information and perform tasks such as description generation and question answering based on visual inputs. These models require extensive pre-training and fine-tuning on large image-text or video-text datasets, where the quality of the descriptions plays a crucial role in performance \cite{chen2023sharegpt4v,bai2024survey}. As a result, ongoing research is focused on how to generate high-quality descriptions more effectively.

% \subsection{Multi-Modal Large Language Models}
\subsection{Image Captioning with MLLMs}

A high-quality image caption should accurately reflect the image content, include all key elements, and remain relevant to the scene. MLLMs have significantly advanced image captioning by generating more detailed and contextually rich descriptions compared to traditional models \cite{vinyals2016show,you2016image}. However, challenges such as hallucinations \cite{bai2024hallucination-survey} and lack of fine-grained detail \cite{pi2024image-textual} persist. Methods like LLaVA-Detailed-Captions \cite{liu2024llava}, ShareGPT4V \cite{chen2023sharegpt4v}, and ALLaVA \cite{chen2024allava} generate high-quality captions through either closed-source models or fine-tuning on specialized datasets. However, these approaches rely on expensive APIs and training resources. Other methods \cite{pi2024image-textual, dong2024capture}, which enhance captioning without retraining to improve accuracy and reduce hallucinations,  but they only enhance global descriptions using object-level information and phrase-level hallucinations. The AnyRes strategy in Llava-onevision \cite{li2024llava} divides images into evenly-sized patches for input to enhance perception. However, this approach does not account for the interactions between objects or the overall semantic content of the image. To address these issues, we propose a train-free \textit{divide-then-aggregate} method with semantic-level filtering to capture finer details, including object interactions within regions, and mitigate hallucinations.

%-------------------------------------------------------------------------

%-------------------------------------------------------------------------

%% file: 3_method.tex
\section{Method}

\subsection{Patch Division for Enhanced Local Perception}
\label{Sec3.2:Patch Slicing}

As analyzed in \cref{sec:intro}, dividing the image into patches improves the model’s local perception. A straightforward method \cite{chen2024poca} to split the image is to divide it into $n$ equal patches, offering a patch perspective:

\begin{equation}
\{R_1, R_2, \ldots, R_n\} = f_e(I),
\end{equation}
where $I$ is the image, $f_e$ represents the equal partitioning function, and $R_i$ denotes the resulting patches. 

However, dividing an image into equal parts can disrupt object relationships, often fragmenting cohesive elements and causing inconsistencies or wrong descriptions. Additionally, key semantic information may span across multiple regions, obscuring important details. To address this, we propose a division approach that integrates both spatial and semantic considerations, as shown in \cref{fig:method} (A), preserving critical object relationships and semantic meaning within each patch.

For spatial patch division, we apply a relation-based method that respects the spatial structure of objects within the image, avoiding the incorrect division of a single object across multiple patches. We begin by dividing the image into four quadrants to anchor spatial context. Then, using an object detection model, we identify potential objects and retrieve proposals through the model’s Region Proposal Network (RPN) and Region of Interest (ROI) head. To limit false positives, we discard proposals with confidence scores below 0.3. For each valid object proposal, we compute the IoU with the quadrants; if the IoU exceeds a threshold, the object is assigned to that patch. This process generates four refined spatial patches $R_i^s \ (\text{where } i \in \{1, 2, 3, 4\})$ using the relation-based method $f_s$, enriching the visual representation while maintaining object integrity.

For semantic patch division, we aim to capture the primary semantic content of the image using a semantics-based approach. This helps narrow the model's perceptual field, allowing it to focus more on the information of key semantic objects. We start by generating a high-level semantic description using a small VLM like BLIPv2 \cite{li2023blipv2}. Then, leveraging a LLM, we identify overlapping objects between this high-level semantic description and the detected objects generated by an object detection model. We combine the bounding boxes of overlapping objects into a new patch $R^t$. Through this semantics-based approach $f_t$, the new patch $R^t$ represents the primary semantic patch of the image. 

Next, by combining spatial and semantic division, we generate five key patches:
\begin{equation}
\{R^s_1, R^s_2, R^s_3, R^s_4, R^t\} = \{f_s(I), f_t(I)\}.
\end{equation}

After obtaining the divided patches, we generate multiple candidate visual descriptions for each patch to capture detailed information and enhance reliability through hierarchical aggregation.

\subsection{Hierarchical Aggregation}
\label{Sec3.3:Hierarchical aggregation}

After dividing the image into patches and performing local perception, we need to aggregate the local information to obtain a detailed global caption. Our hierarchical aggregation process includes two stages: (1) Intra-patch aggregation (\cref{Sec3.3.1:Intra-patch Aggregation}): Fusing multiple candidate descriptions within a single patch into a reliable description. (2) Inter-patch aggregation (\cref{Sec3.3.2:Inter-patch Aggregation}): Combining semantic and spatial patch descriptions into a comprehensive global description. During the hierarchical aggregation process, we employ a semantic-level filtering strategy (\cref{Sec3.3.3:Semantic-level Filtering Strategy}) to reduce hallucinations and description conflicts.

\subsubsection{Intra-patch Aggregation}
\label{Sec3.3.1:Intra-patch Aggregation}

In the Intra-patch aggregation stage, each patch has multiple candidate descriptions that need to be merged into a coherent and complete description. Among these descriptions, some sentences provide consistent information about the same object, indicating high semantic certainty. However, there may also be contradictory descriptions about the same object, as well as descriptions that appear only once across the candidates, which may indicate lower certainty or potential errors. To address this, we apply a semantic filtering strategy (\cref{Sec3.3.3:Semantic-level Filtering Strategy}) to classify the candidate descriptions into three categories—\textit{Same}, \textit{Contradictory}, and \textit{Unique}—based on their semantic analysis. A reliability list is then generated, which helps to merge the descriptions into a unified and accurate patch description.

\subsubsection{Inter-patch Aggregation}
\label{Sec3.3.2:Inter-patch Aggregation}

In the Inter-patch aggregation stage, as shown in \cref{fig:method} (C), we aggregate the semantic patch descriptions into the global description.  We calculate the IoU of the semantic patch with the global image. If the IoU falls below a predefined threshold, the patch’s information is considered supplemental. In such cases, we prompt the LLM to update the global description by incorporating additional details and correcting any inaccuracies based on the primary semantic information. Next, we process the spatial patch descriptions by calculating the IoU between adjacent patch pairs (4 pairs in total) to assess their overlap. If the overlap exceeds a threshold, we merge the corresponding descriptions to reduce redundancy. This step also applies semantic filtering to ensure consistency and accuracy. Finally, after merging the patch descriptions, we integrate the refined descriptions with the global description using the LLM, adding further details and corrections to generate a complete and polished global description. 
\subsubsection{Semantic Filtering Strategy }
\label{Sec3.3.3:Semantic-level Filtering Strategy}

In the Intra-patch aggregation stage, each patch generates multiple candidate descriptions. In the subsequent Inter-patch aggregation stage, when the IoU between two spatial patches is large, it indicates a high overlap in content, suggesting that these patches represent similar image regions. Therefore, we apply a \textit{semantic filtering strategy} to merge these descriptions into a coherent and reliable representation that retains fine-grained details.

Based on the analysis in \cref{Sec3.3.1:Intra-patch Aggregation}, we perform semantic analysis on the candidate descriptions. We use a LLM with strong semantic understanding to evaluate and compare the candidate descriptions. Specifically, we first check whether multiple sentences from different candidates describe the same object, which indicates high certainty and suggests that the description is likely correct \cite{DBLP:conf/iclr/XiongHLLFHH24}. Next, we analyze whether any descriptions contradict each other, which helps mitigate hallucinations in the descriptions. Finally, to enhance fine-grained details, we instruct the LLM to identify sentences that describe an object but appear only once across the candidates, as these are likely to indicate either hallucinations or missing information. Based on this analysis, the candidate descriptions are categorized into three types: \textit{Same}, \textit{Contradictory}, and \textit{Unique}.

For \textit{Same} category, we obtain a high-certainty sentence list, denoted as \(\{x^1_s, \ldots, x^p_s\}\), as shown in \cref{fig:method} (B). This consolidated description includes key visual info like objects, attributes, and relationships, ensuring a more accurate and reliable global description with fewer hallucinations.

For \textit{Contradictory} category, we maintain a list of conflicting pairs. For example, as shown in \cref{fig:method} (B), contradictions are flagged to prevent hallucinations. We use the Blip2Score \cite{bianco2023blip2score-testing} to measure the similarity between contradictory sentences and the image. The Image-Text Matching similarity \( \text{Sim}(R, y) \) and the Image-Text Matching score \( M(R, y) \) are averaged as follows:
\begin{equation}
\text{Blip2Score} = \frac{\text{Sim}(R, y) + M(R, y)}{2}.
\end{equation}
If one of the contradictory sentences has a higher Blip2Score and exceeds a set threshold, it is retained as the correct description sentence, denoted as \(\{x^1_c, \ldots, x^p_c\}\). 

For \textit{Unique} category, we identify descriptions containing fine-grained details that may either be correct or potentially erroneous, as shown in \cref{fig:method} (B). These descriptions require further evaluation to determine whether they should be included as supplementary information. We calculate the Blip2Score for each unique sentence, retaining those exceeding the threshold, denoted as \(\{x^1_u, \ldots, x^m_u\}\).

Finally, we combine sentences from the \textit{Same}, \textit{Contradictory}, and \textit{Unique} categories into a set:
\begin{equation}
X_{\text{supp}} = \{x^1_s, \ldots, x^p_s, x^1_c, \ldots, x^q_c, x^1_u, \ldots, x^m_u\}.
\end{equation}
This set of high-certainty sentences, along with the initial candidate descriptions, is then used by the LLM to generate a refined, comprehensive description for both Intra-patch aggregation and Inter-patch aggregation. This process effectively reduces hallucinations and enhances the quality and reliability of the image captions. The details of prompts are provided in the appendix.

%% file: 4_experiment.tex
\begin{table*}[ht]
        \centering
    \small
    \renewcommand{\arraystretch}{0.6}
     \setlength{\tabcolsep}{5.2pt}
     \setlength{\extrarowheight}{0pt}
    \begin{tabular}{l|c|cccc|ccccc} % 将 tabular* 替换为 tabular
        \toprule
        GroundTruth & Method & BLEU-1 & BLEU-2 & BLEU-3 & BLEU-4 & CIDEr & METEOR & ROUGE & SPICE & WMD \\
        \midrule
        \rowcolor{gray!20}
        \multicolumn{11}{c}{\textit{\textbf{LLaVA-1.5-7B}}} \\
        \multirow{5}{*}{GT-\{LLaVA\}} & Global & 9.16  & 5.70 & 3.37 & 2.07 & 0.47 & 10.62 & 19.54 & 17.65& 41.37 \\
         & IT \cite{pi2024image-textual} & 25.36 &15.65 &	9.31 &	5.76 &	0.73 &	12.82 	&20.16 &	19.06 	&42.07  \\ 
         & PoCa \cite{chen2024poca} & 0.59 &	0.34 &	0.18 &	0.10 &	0.00 &	5.95 &	11.47 &	13.63 &	36.61  \\
         & Synthesized \cite{dong2024capture} & 20.65 &	12.05 	&6.72& 	3.93 &	0.96& 	14.00 &	20.95 	&19.67 &	43.04  \\
          & Ours & \textbf{34.06} & \textbf{19.09} & \textbf{10.35} & \textbf{5.85} & \textbf{5.17} & \textbf{17.73} & \textbf{21.01} & \textbf{20.38} & \textbf{43.74} \\
        \midrule
        \multirow{5}{*}{GT-\{GPT4-V\}} & Global & 7.50 &	4.05 &	1.94 &	0.98 	&0.00 &	9.11 &	15.67 &	13.29 &	37.74 \\
         & IT \cite{pi2024image-textual}& 19.19 &	10.53 &	5.28 &	2.77 &	0.56 &	11.23 &	16.61 &	14.87 &	38.65 \\
         & PoCa \cite{chen2024poca}& 0.36 &	0.18 &	0.09 &	0.04 &	0.00 &	5.03 &	9.60 &	10.14 &	34.23  \\
         & Synthesized \cite{dong2024capture}& 18.78 	&10.16& 	5.21 &	2.83 &	0.24 &	12.75& 	18.51 &	16.75& 	40.29 \\
          & Ours & \textbf{29.36} &	\textbf{15.12} &	\textbf{7.27}& 	\textbf{3.61} &	\textbf{2.04} &	\textbf{15.67} &	\textbf{18.64} &	\textbf{17.25} &	\textbf{40.62}   \\
        \rowcolor{gray!20}
        \multicolumn{11}{c}{\textit{\textbf{LLaVA-1.6-7B}}} \\
         \multirow{5}{*}{GT-\{LLaVA\}} & Global & 23.36 & 13.54 & 7.84 & 4.70 & 1.10 & 14.74 & 21.16 & 20.25& 43.75 \\
         & IT \cite{pi2024image-textual}& 32.45	&19.02	&11.07	&6.62	&2.31	&16.70	&21.57	&21.38	&44.35\\
         & PoCa \cite{chen2024poca}& 2.26	&1.32	&0.73	&0.41	&0.39	&7.74	&13.64	&15.85	&38.89 \\
         & Synthesized \cite{dong2024capture}&28.43 &	16.18 &	9.10 &	5.29 &	1.75 	&16.11 	&21.67 &	20.82 	&44.30  \\
          & Ours & \textbf{42.84} & \textbf{24.08} & \textbf{13.22} & \textbf{7.48} & \textbf{5.50} & \textbf{20.71} & \textbf{21.69} & \textbf{21.78} & \textbf{45.95} \\
        \midrule
        \multirow{5}{*}{GT-\{GPT4-V\}} & Global & 19.98&	10.63	&5.46&	3.00	&0.39	&13.61	&18.56	&19.32	&40.74 \\
         & IT \cite{pi2024image-textual}& 30.38	&16.47	&8.63	&4.87	&0.51	&15.18	&19.16	&20.14	&41.32\\
         & PoCa \cite{chen2024poca}& 3.09	&1.67	&0.86	&0.48	&0.00	&8.13	&13.63	&16.23	&37.89 \\
         & Synthesized \cite{dong2024capture} &28.79 &	15.69& 	8.42 &	4.81 &	1.47 &	15.99& 	20.22& 	20.45& 	42.50  \\
          & Ours & \textbf{37.30} & \textbf{20.07} & \textbf{10.62} & \textbf{6.00} & \textbf{3.60} & \textbf{18.67} & \textbf{20.39} & \textbf{20.99} & \textbf{43.33} \\
          \rowcolor{gray!20}
        \multicolumn{11}{c}{\textit{\textbf{Mini-Gemini-7B}}} \\
        \multirow{5}{*}{GT-\{LLaVA\}} & Global &27.54 &	15.48 &	8.36 &	4.77 	&4.74 	&16.04 &	21.35& 	20.79 &	43.63  \\
         & IT \cite{pi2024image-textual}& 32.24 	&18.21 &	9.91 &	5.65 &	6.82 &	16.98 &21.37 	&21.34 	&43.96  \\
         & PoCa \cite{chen2024poca} & 1.74 &	0.97 &	0.52 &	0.29 &	0.00 &	7.13 	&13.03 	&15.38 	&38.08  \\
         & Synthesized \cite{dong2024capture} & 35.80 &	19.79 &	10.53 &	5.85 &	4.13& 	18.00 &	21.02 &	20.81 &	44.06   \\
          & Ours & \textbf{42.36} & \textbf{23.76}& \textbf{12.79} & \textbf{7.17} & \textbf{13.43} & \textbf{21.37}  & \textbf{21.43} & \textbf{21.37} & \textbf{45.09} \\
        \midrule
        \multirow{5}{*}{GT-\{GPT4-V\}} & Global &30.18 	&17.96 	&11.09 	&7.34 	&2.54 	&17.18 	&\textbf{23.51} 	&22.62 	&44.09 \\
         & IT \cite{pi2024image-textual}& 33.53 &	19.99 	&12.34 	&8.12 &	2.53 &	18.08 &	23.35 &	22.71 &	44.28   \\
         & PoCa \cite{chen2024poca} & 2.65 	&1.55 	&0.92 	&0.56 	&0.06 	&8.19 	&14.26 	&16.85 	&38.70  \\
         & Synthesized \cite{dong2024capture} & 37.40 &	21.54 	&12.91 &	8.30 &	1.60 &	18.68 &	22.81& 	22.58 &	44.36   \\
          & Ours & \textbf{40.30} & \textbf{23.51} & \textbf{13.65} & \textbf{8.37} & \textbf{2.66} & \textbf{22.49} & 22.68 & \textbf{22.99} & \textbf{46.08} \\
        \bottomrule
    \end{tabular}
    \normalsize 
    \caption{Evaluation of image descriptions on DID-Bench using open-source models, GT-\{LLaVA\} and GT-\{GPT4-V\} refer to ground truth captions generated by LLaVA and GPT-4Vision, respectively. The increase in BLEU scores and ROUGE score indicates that the generated captions are structurally and lexically closer to the reference captions. Improvements in CIDEr, METEOR, SPICE, and WMD scores reflect enhanced semantic alignment with the content of the reference captions. }
    \label{tab:table1}
\end{table*}
\begin{table*}[ht]
    \centering
    \small
     \setlength{\tabcolsep}{5.2pt}
    \renewcommand{\arraystretch}{0.6}
    \begin{tabular}{l|c|cccc|ccccc} % 将 tabular* 替换为 tabular
        \toprule
        GroundTruth & Method & BLEU-1 & BLEU-2 & BLEU-3 & BLEU-4 & CIDEr & METEOR & ROUGE & SPICE & WMD \\
        \midrule
        \rowcolor{gray!20}
        \multicolumn{11}{c}{\textit{\textbf{GLM-4V-Plus}}} \\
        \multirow{5}{*}{GT-\{LLaVA\}} & Global  & 10.46  & 6.31   & 3.79   & 2.40   & 1.28  & 11.46  & 19.08 & 19.46 & 42.21   \\
        & IT \cite{pi2024image-textual} & 23.69  & 14.38  & 8.63   & 5.45   & 1.20  & 14.18  & 20.65 & 21.07 & 43.26\\
         & PoCa \cite{chen2024poca} & 2.03   & 1.19   & 0.67   & 0.40   & 0.00  & 7.72   & 13.44 & 16.62 & 39.11 \\
         & Synthesized \cite{dong2024capture} & 24.74  & 14.27  & 8.23   & 4.90   & 1.26  & 15.29  & 21.35 & 21.53 & 44.04  \\
          & Ours & \textbf{34.99} & \textbf{20.07} & \textbf{11.40} & \textbf{6.76} & \textbf{1.36} & \textbf{18.19} & \textbf{21.43} & \textbf{22.24} & \textbf{45.18} \\
        \midrule
        \multirow{5}{*}{GT-\{GPT4-V\}} & Global &7.70 &	4.12 &	2.15 &	1.21& 	0.42 &	9.90 	&15.52& 	17.34& 	39.05  \\
        & IT \cite{pi2024image-textual} & 21.35  & 11.65  & 6.24   & 3.54   & 0.03  & 11.97  & 17.04 & 18.58 & 39.91\\
         & PoCa \cite{chen2024poca}  & 1.57   & 0.83   & 0.43   & 0.24   & 0.00  & 7.07   & 12.32 & 14.71 & 37.12 \\
         & Synthesized \cite{dong2024capture} & 20.52  & 11.02  & 5.71   & 3.12   & 0.13  & 13.92  & 18.94 & 20.10 & 41.78  \\
          & Ours & \textbf{30.96} & \textbf{16.68} & \textbf{8.73} & \textbf{4.81} & \textbf{2.76} & \textbf{16.73} & \textbf{19.35} & \textbf{20.24} & \textbf{42.47} \\
        \rowcolor{gray!20}
        \multicolumn{11}{c}{\textit{\textbf{GPT-4o}}} \\
        \multirow{5}{*}{GT-\{LLaVA\}} & Global &8.43 &	5.01 &	2.99& 	1.92& 	1.03 &	11.13& 	17.69 &	20.28& 	42.62  \\
        & IT \cite{pi2024image-textual}& 14.76  & 8.89   & 5.31   & 3.39   & 0.75  & 13.20  & 19.57 & 21.95 & 43.73\\
         & PoCa \cite{chen2024poca}  & 0.66   & 0.39   & 0.22   & 0.14   & 0.00  & 6.57   & 11.90 & 16.01 & 37.94 \\
         & Synthesized \cite{dong2024capture} & 18.41  & 10.50  & 5.93   & 3.46   & 0.00  & 14.18  & 20.08 & 22.44 & 44.14  \\
          & Ours & \textbf{27.06} & \textbf{15.43} & \textbf{8.75} & \textbf{5.19} & \textbf{2.12} & \textbf{16.52} & \textbf{20.12} & \textbf{22.64} & \textbf{44.91} \\
        \midrule
        \multirow{5}{*}{GT-\{GPT4-V\}} & Global & 7.01 	&3.72 &	1.98 &	1.10 	&0.04 &	9.95 &	14.58 &	17.80 &	39.44  \\
        & IT \cite{pi2024image-textual} & 16.46  & 9.03   & 4.92   & 2.83   & 0.02  & 12.89  & 17.28 & 19.26 & 40.59\\
         & PoCa \cite{chen2024poca} & 0.58   & 0.32   & 0.17   & 0.09   & 0.00  & 6.00   & 11.00 & 14.05 & 36.16 \\
         & Synthesized \cite{dong2024capture}  & 15.77  & 8.42   & 4.45   & 2.45   & 0.12  & 12.96  & 17.67 & 19.32 & 41.86  \\
          & Ours & \textbf{22.82} & \textbf{11.89} & \textbf{6.10} & \textbf{3.24} & \textbf{2.11} & \textbf{14.67} & \textbf{17.82} & \textbf{19.57} & \textbf{41.87} \\
        \rowcolor{gray!20}
        \multicolumn{11}{c}{\textit{\textbf{Claude-3.5-Sonnet}}} \\
        \multirow{5}{*}{GT-\{LLaVA\}}&Global & 26.37 &	14.11 &	7.72 	&4.45 &	4.04 &	15.47 &	17.79 &	17.41 &	41.67  \\
        & IT \cite{pi2024image-textual} & 37.04  & 20.38  & 11.38  & 6.67   & 4.10  & 19.04  & 19.96 & 20.40 & 45.01 \\
         & PoCa \cite{chen2024poca} & 4.29   & 2.36   & 1.26   & 0.71   & 0.00  & 8.79   & 13.98 & 16.28 & 39.50 \\
         & Synthesized \cite{dong2024capture} & 30.40  & 16.58  & 8.99   & 5.14   & 2.20  & 17.31  & 20.36 & 20.29 & 44.53  \\
          & Ours & \textbf{39.93} & \textbf{22.14} & \textbf{11.87} & \textbf{6.65} & \textbf{4.27} & \textbf{20.78} & \textbf{20.69} & \textbf{20.65} & \textbf{45.06} \\
        \midrule
        \multirow{5}{*}{GT-\{GPT4-V\}} & Global & 19.52 	&9.58 &	4.76 &	2.54 	&2.09 	&12.81 	&14.81 &	15.02 &	38.89  \\
        & IT \cite{pi2024image-textual} & 33.73  & 17.39  & 8.92   & 4.81   & 1.87  & 17.50  & 19.09 & 19.46 & 43.04\\
         & PoCa \cite{chen2024poca}  & 4.43   & 2.30   & 1.15   & 0.62   & 0.00  & 8.56   & 13.19 & 15.09 & 38.35 \\
         & Synthesized \cite{dong2024capture} & 27.70  & 14.29  & 7.18   & 3.79   & 1.12  & 16.04  & 19.07 & 19.45 & 43.01  \\
          & Ours & \textbf{36.83} & \textbf{19.00} & \textbf{9.58} & \textbf{5.11} & \textbf{2.75} & \textbf{20.01} & \textbf{19.49} & \textbf{19.67} & \textbf{43.46} \\
        \bottomrule
    \end{tabular}
    \normalsize 
    \caption{Evaluation of image descriptions on DID-Bench using closed-source models, with GT-\{LLaVA\} and GT-\{GPT4-V\} as ground truth captions generated by LLaVA and GPT-4Vision, respectively.  Our method can also improve caption quality for closed-source models.}
    \label{tab:table2}
\end{table*}
\section{Experiment}

\subsection{Setup}
\subsubsection{Evaluation Metrics and Benchmarks.}
% \textbf{Reference-Based Metircs.}
\label{4.1.1}
% \textbf{Becnchmarks and Evaluation Metrics.} 
We evaluated our method on various benchmarks like DID-Bench \cite{pi2024image-textual}, D2I-Bench \cite{pi2024image-textual}, DetailCaps Benchmark \cite{dong2024capture}.

\noindent \textbf{DID-Bench.} DID-Bench includes 200 images extracted from COCO \cite{lin2014COCO-data}, with 100 labeled by LLaVA and 100 by GPT-4Vision, referred to as GT-\{LLaVA\} and GT-\{GPT4-V\}, respectively. These annotations were manually refined to correct any missing or erroneous visual elements.  We adopt metrics such as BLEU \cite{papineni2002bleu}, CIDEr \cite{vedantam2015cider}, METEOR \cite{lavie-agarwal-2007-meteor}, ROUGE-L \cite{lin2004rouge}, SPICE \cite{anderson2016spice}, and WMD \cite{DBLP:conf/eacl/KilickayaEIE17} to evaluate the quality of the generated descriptions. Additionally, we incorporate CLIP-S \cite{hessel2021clip_score}, Polos \cite{wada2024polos}, and CHAIR \cite{rohrbach2018chair_metric} metrics to assess caption-image alignment and hallucination reduction. These metric details are in the appendix.

\noindent \textbf{D2I-Bench.} D2I-Bench calculates CLIP-score and DINO-score to assess performance by generating descriptions for image reconstruction. The similarity between reconstructed and original images reflects the process's effectiveness. The image generation model used is Pixart-$\alpha$ \cite{DBLP:conf/iclr/ChenYGYXWK0LL24}.

\noindent \textbf{DetailCaps Benchmark.} The CAPTURE metric, introduced within the framework of the detailcaps benchmark \cite{dong2024capture}, utilizes scene graph parsing to align objects, attributes, and relationships with reference captions for more human-like assessment. This benchmark uses images from COCO \cite{lin2014COCO-data}, LAION \cite{schuhmann2021LAION}, SAM \cite{kirillov2023SAM}, SBU\cite{ordonez2011SBU-dataset}, Flickr \cite{young2014Flickr-dataset}, Coyo \cite{kakaobrain2022Coyo-700m}, and CC \cite{sharma2018CC-dataset}, with text descriptions generated by GPT-4Vision. We randomly selected 200 images for evaluation.

\subsubsection{Baselines and Models.}
\textbf{Baselines.} To evaluate our method, we chose three training-free image caption enhancement methods—IT \cite{pi2024image-textual}, PoCa \cite{chen2024poca}, and Synthesized \cite{dong2024capture}. IT and Synthesized use extra visual experts to reduce hallucinations and improve captions, while PoCa uses a quadrant-based approach to integrate localized image information. We also compared these methods with two fine-tuned models, ShareCaptioner\cite{chen2023sharegpt4v} and ALLaVA \cite{chen2024allava}, both of which use GPT-4Vision-generated detailed image captions to enhance captioning effectiveness. ``Global" refers to captions from the base model.

\noindent \textbf{Models.} We conducted evaluations on open-source models, including LLaVA-1.5 (7B), LLaVA-1.6 (7B) \cite{liu2024llava}, and Mini-Gemini (7B) \cite{li2024mini-gemini}, as well as on close-source models such as GPT-4o \cite{hurst2024gpto}, Claude-3.5-Sonnet \cite{claude3.5}, and GLM-4V-Plus \cite{glm2024chatglm}. For fine-tuned models, since ShareCaptioner is based on LLaVA 1.5, we use LLaVA 1.5 with our method for comparison. All methods that required LLM integration were implemented using LLaMA 3.1 (8B) \cite{dubey2024llama}.

\begin{table*}[ht]
    \centering
    \small
      \setlength{\tabcolsep}{4.8pt}
          \renewcommand{\arraystretch}{0.6}
  \setlength{\belowcaptionskip}{-4mm}
    \begin{tabular}{l|c|cccc|ccccc} % 将 tabular* 替换为 tabular
        \toprule
        GroundTruth & Method & BLEU-1 & BLEU-2 & BLEU-3 & BLEU-4 & CIDEr & METEOR & ROUGE & SPICE & WMD \\
        \midrule
        \multirow{4}{*}{GT-\{LLaVA\}} & LLaVA-1.5  &9.16 	&5.70 &	3.37 &	2.07 &	0.47 	&10.62& 	19.54 &	17.65 	&41.37   \\
        & ShareCaptioner \cite{chen2023sharegpt4v}  &31.15 &	17.16 	&9.15 &	5.20 &	2.30 &	16.71 &	20.80 &	20.21 &	42.51 \\
         & ALLaVA \cite{chen2024allava} &29.77 	&15.62 &	8.07 	&4.34& 	2.59 &	15.38 &	19.62 	&16.76 	&40.92  \\
          & LLaVA-1.5+Ours & \textbf{34.06} & \textbf{19.09} & \textbf{10.35} & \textbf{5.85} & \textbf{5.17} & \textbf{17.73} & \textbf{21.01} & \textbf{20.38} & \textbf{43.74} \\
        \midrule
        \multirow{4}{*}{GT-\{GPT4-V\}} & LLaVA-1.5 &7.50 	&4.05 	&1.94 &	0.98 &	0.00 	&9.11 	&15.67 	&13.29& 	37.74   \\
        & ShareCaptioner \cite{chen2023sharegpt4v}  & \textbf{33.42} &	\textbf{20.11} &	\textbf{12.66} &	\textbf{8.68} &	\textbf{2.69} &	\textbf{18.17} &	\textbf{24.46}& 	\textbf{24.85} 	&\textbf{44.13} \\
         & ALLaVA \cite{chen2024allava} &26.27 &	13.01 &	6.15 &	3.18 &	2.16 &	13.84 &	18.06 &	14.89 	&37.67  \\
          & LLaVA-1.5+Ours & 29.36 &	15.12 &	7.27 	&3.61 	&2.04& 	15.67 &	18.64 &	17.25 	&40.62  \\
        \bottomrule
    \end{tabular}
    \normalsize 
    \caption{Evaluation of image descriptions on DID-Bench, GT-\{LLaVA\} and GT-\{GPT4-V\} refer to ground truth captions generated by LLaVA and GPT-4Vision, respectively. Compared to other training-based models, our approach also achieves strong performance.}
    \label{tab:table3}
\end{table*}

\subsubsection{Experiment Details.}
We employed OVDet \cite{wu2023ovdet} to provide region proposals. For concise image caption generation (\cref{Sec3.2:Patch Slicing}), we used BLIPv2 (2.7B) \cite{li2023blipv2} as the small VLM. We set the Blip2Score threshold to 0.3 and the IoU threshold to 0.4. To generate descriptions for images and regions, we utilized prompts such as \textit{Describe this image in detail}. Specifically, for DID-Bench \cite{pi2024image-textual}, we used the official prompts to generate captions. We use the same prompt to generate spatial patch, semantic patch, and global descriptions, with the temperature set to 0.7 for the multiple descriptions. For semantic filtering and aggregation, we employed LLaMA 3.1 (8B) \cite{dubey2024llama} as our LLM. Additional experimental content, details, and prompts can be found in the appendix.

\subsection{Quantitative Results}
\textbf{DID-Bench.}
\cref{tab:table1} shows results comparing our method to other training-free approaches on open-source models. Our method consistently outperforms others across various MLLMs, achieving significant improvements (up to +7.11 on METEOR and +4.7 on CIDEr). Even when baseline models perform well, our approach further boosts their performance, unlike other methods that may degrade CIDEr scores. This advantage likely arises from our patch enhancing local perception. Furthermore, as shown in \cref{tab:table2}, our method also significantly improves performance on closed-source models. This demonstrates that our method performs well on both open-source and closed-source models.

We also compared our method with fine-tuned models, including ShareCaptioner \cite{chen2023sharegpt4v} and ALLaVA \cite{chen2024allava}, both trained on extensive, high-quality image-text datasets. As shown in \cref{tab:table3}, while ShareCaptioner excels on GT-\{GPT4-V\} due to its fine-tuning on GPT4-Vision data, our method achieved the best performance on GT-\{LLaVA\} and outperforms ALLaVA, demonstrating that our approach is competitive with fine-tuned models.

\begin{table}[h]
    \centering
    \small
        \renewcommand{\arraystretch}{0.6}
    \setlength{\belowcaptionskip}{-2mm}
    \begin{tabular}{l |cc}
        \toprule
        Description & CLIP-score & DINO-score \\
        \midrule
         LLaVA1.5 &73.09 &	63.80    \\
         LLaVA1.5+IT \cite{pi2024image-textual}&73.09 &	64.24     \\
        LLaVA1.5+PoCa \cite{chen2024poca}&72.04 &	62.61  \\
         LLaVA1.5+Syn \cite{dong2024capture}&71.71&63.63 \\
         LLaVA1.5+Ours & \textbf{73.67} 	&\textbf{66.32}  \\
        \midrule
         LLaVA1.6 & 76.33 &	68.39   \\
         LLaVA1.6+IT \cite{pi2024image-textual}&76.18 &	68.19   \\
         LLaVA1.6+PoCa \cite{chen2024poca}&74.39 &	66.59   \\
         LLaVA1.6+Syn \cite{dong2024capture}& 75.27&67.69  \\
         LLaVA1.6+Ours & \textbf{76.48} &	\textbf{69.85} 	  \\
         \midrule
         Mini-Gemini &76.02 &	68.15   \\
         Mini-Gemini+IT \cite{pi2024image-textual}& \textbf{76.21} 	&68.18    \\
         Mini-Gemini+PoCa \cite{chen2024poca}&75.20 	&66.86    \\
         Mini-Gemini+Syn \cite{dong2024capture}& 75.31 &67.84  \\
         Mini-Gemini+Ours & 76.20 &\textbf{68.48}  \\
        \bottomrule
    \end{tabular}
    \normalsize 
    \caption{D2I-Bench Results. The images we generate achieve strong similarity scores compared to the original images.}
    \label{tab:table4}
\end{table}

% \textbf{D2I-Bench and LIN-Bench.}
\noindent \textbf{D2I-Bench.} 
As shown in \cref{tab:table4}, our method’s captions effectively capture visual details, producing reconstructed images closely aligned with originals. Qualitative examples can also be found in \cref{fig:case_d2i}, where our method captures fine-grained features such as whether a person is wearing a hat and the positional relation between a woman and a horse.

\begin{table}[t]
    \centering
    \small
        \renewcommand{\arraystretch}{0.6}
  \setlength{\belowcaptionskip}{-4mm}
     \setlength{\tabcolsep}{3pt}
    \begin{tabular}{l |cccc}
        \toprule
         Description & CAPTURE & $F1_{obj}$ & $F1_{attr}$ & $F1_{rel}$ \\
        \midrule
         LLaVA1.5 &49.99 & 55.76 & 44.43 & 49.48   \\
         LLaVA1.5+IT \cite{pi2024image-textual} &51.98& 	56.37 &	48.22 &	50.44   \\
         LLaVA1.5+PoCa \cite{chen2024poca} & 47.38 &	54.57 	&40.64 	&46.28  \\
         LLaVA1.5+Syn \cite{dong2024capture} & 57.51 & 62.06 & 55.54 &	51.07   \\
         LLaVA1.5+Ours & \textbf{58.05} & \textbf{62.23} & \textbf{56.09} & \textbf{52.51}  \\
        \midrule
         LLaVA1.6 & 61.03 &	64.59 &	60.39 &	53.73   \\
         LLaVA1.6+IT \cite{pi2024image-textual} & 61.21 	&64.76 &	60.68& 	53.68   \\
         LLaVA1.6+PoCa \cite{chen2024poca} &55.38 &	58.61 &	54.47& 	49.61  \\
         LLaVA1.6+Syn \cite{dong2024capture} & 61.16 	&65.17 &	60.24 &	53.44   \\
         LLaVA1.6+Ours & \textbf{63.51} & \textbf{66.95} & \textbf{63.15} & \textbf{55.80}  \\
         \midrule
         Mini-Gemini & 62.37 	&66.77 	&61.15 	&54.40   \\
         Mini-Gemini+IT \cite{pi2024image-textual} & 62.53 	&66.85 	&61.40 &	54.54  \\
         Mini-Gemini+PoCa \cite{chen2024poca} & 55.08 &	59.73 &	52.68 &	49.44  \\
         Mini-Gemini+Syn \cite{dong2024capture} & 62.53 &	66.68 &	61.49 &	54.77   \\
         Mini-Gemini+Ours & \textbf{64.49} & \textbf{68.15} & \textbf{63.92} & \textbf{56.75}  \\
        \bottomrule
    \end{tabular}
    \normalsize 
    \caption{CAPTURE scores on DetailCaps benchmark. Higher object, attribute, and relation F1 scores mean better performance in capturing these aspects.}
    \label{tab:table7}
\end{table}

\noindent \textbf{DetailCaps Benchmark.}

Results in \cref{tab:table7} show that our method achieves top performance, effectively capturing object and attribute details. Additionally, it improves the relational F1 score, demonstrating the benefit of region-based descriptions for richer, more accurate captions. Even with advanced MLLMs, our method shows significant gains in \(F1_{rel}\) (+0.37 vs. +2.35 in Mini-Gemini). The CAPTURE score on DID-Bench is computed, results in appendix.

\noindent \textbf{Learnable Multimodal and Hallucination Metrics.}
We used CLIP-S, Polos, and CHAIR metrics to evaluate caption-image alignment and hallucination reduction in DID-Bench. As shown in the \cref{tab:clips}, our approach improves image content reflection and significantly reduces hallucinations. For GPT-4o, the CHAIRs drops by 2.76 compared to the global baseline.  Additionally, the difference in hallucinations with and without semantic filtering is significant, with hallucinations dropping from 30.2\% to 8.42\%.
 
\begin{table}[b]
   \centering
    \small
        \renewcommand{\arraystretch}{0.6}
  \setlength{\belowcaptionskip}{-4mm}
     \setlength{\tabcolsep}{1.1pt}
    \begin{tabular}{c|c|cccc}
   \toprule
        Model & Method & CLIP-S  $\uparrow$ & Polos  $\uparrow$ &CHAIRs $\downarrow$ & CHAIRi $\downarrow$   \\
         \midrule
\multirow{6}{*}{GPT-4o} & Global & 40.59& 76.35  & 11.18& 5.59 \\
& IT & 41.00& 76.61  & 25.25& 5.54 \\
& Synthesized & 41.13& 76.87  & 29.71& 7.09 \\
& PoCa & 38.36& 72.47  & 18.32& 7.55 \\
&w/o. semantic&41.10 & 75.53& 30.20 &	8.54 \\
&Ours & \textbf{41.19 } & \textbf{76.98}& \textbf{8.42} & \textbf{4.30} \\
  \midrule
    \multirow{6}{*}{LLaVA-1.5} & Global & 40.59  & 71.42& 52.48 	&17.41 \\
    &IT &  40.71 &71.82 &  52.48 &	15.38\\
    &Synthesized & 40.77  & 71.89& 53.47 &	15.44 \\
    &PoCa & 37.26 & 66.64& 55.45 &	27.50 \\
    &w/o. semantic&40.65 & 71.30& 81.67 &	23.89 \\
    &Ours & \textbf{40.78 } & \textbf{72.81}& \textbf{49.50 } & \textbf{15.33} \\
 % \midrule
 %  \multirow{6}{*}{LLaVA-1.6} & Global & 41.31  & 75.87& 27.72 &	6.96 \\
 %    &IT &41.32  & 76.01&30.69 &	\textbf{5.70}  \\
 %    &Synthesized & 41.16  & 76.38& 35.64 &	8.70  \\
 %    &PoCa & 39.00 & 72.72& 21.29 &	7.76 \\
 %    &w/o. semantic filter&41.11 & 76.60& 50.99 &	11.19 \\
 %    &Ours & \textbf{41.34 } & \textbf{76.82}& \textbf{23.76} 	&8.34  \\
 %   \hline
 % \multirow{5}{*}{Mini-Gemini} & Global & 40.88 & 76.67& 34.65 	&10.95  \\
 %    &IT & 40.85 & 76.60  & 37.62 &	10.18 \\
 %    &Synthesized & 40.70& 76.55&38.12 	&9.92   \\
 %    &PoCa & 38.81 & 72.71 & \textbf{28.22} 	&11.92\\
 %    &Ours & \textbf{40.91} & \textbf{76.72} & \underline{28.71} & \textbf{9.47} \\
   \bottomrule
\end{tabular}
\caption{Multimodal Metrics: CLIP-S, Polos, and Hallucination Metric CHAIR on DID-Bench.}
\label{tab:clips}
\end{table}

\begin{figure*}[ht]
  \centering
  \vspace{2pt}
  
    \setlength{\abovecaptionskip}{0.1cm}
  \setlength{\belowcaptionskip}{-2mm}
   \includegraphics[width=\linewidth]{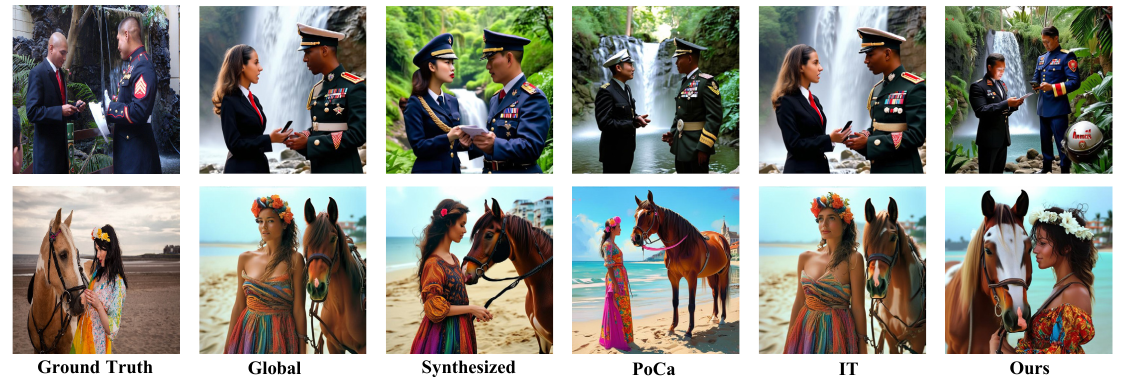}
    \caption{
The visualization of D2I-Bench shows that our method is able to capture more image details, as well as object attributes and relationships, resulting in generated images that are more similar to the original images.}
    \label{fig:case_d2i}
\end{figure*}
% \begin{figure}[h]
%   \centering
%   % \fbox{\rule{0pt}{2in} \rule{0.9\linewidth}{0pt}}
%    \includegraphics[width=1\linewidth]{sec/vqa.pdf}
    
%    \caption{The performance of different methods and models on the VQA task shows that our method achieves the best results.}
%    \label{fig:vqa}
% \end{figure}
\begin{table*}[ht]
    \centering
    \small
     \setlength{\belowcaptionskip}{-3mm}
        \renewcommand{\arraystretch}{0.6}
      \setlength{\tabcolsep}{3.3pt}
    \begin{tabular}{l|c|cccc|ccccc}
        \toprule
        GroundTruth & Method & BLEU-1 & BLEU-2 & BLEU-3 & BLEU-4 & CIDEr & METEOR & ROUGE & SPICE & WMD \\
        \midrule
\multirow{5}{*}{GT-\{LLaVA\}} & w/o. semantic patch & 37.82 & 21.36 & 11.91 & 6.92 & 4.47 & 19.05 & 21.21 & 21.60 & 45.36 \\ 
& w/. four equal           & 40.08 & 22.57 & 12.29 & 6.90 & 4.76 & 19.50 & 21.21 & 21.77 & 45.51 \\ 
& w/o. semantic filtering  & 41.99 & 23.63 & 13.09 & \textbf{7.61} & 2.93 & 20.18 & 21.44 & 21.73 & 45.70 \\
& w/o. hierarchical aggregation    & 15.81 & 8.95  & 5.08  & 3.00 & 1.00 & 12.83 & 18.56 & 19.47 & 42.90 \\ 
& Ours               & \textbf{42.84} & \textbf{24.08} & \textbf{13.22} & 7.48 & \textbf{5.50} & \textbf{20.71} & \textbf{21.69} & \textbf{21.78} & \textbf{45.95} \\
\midrule
\multirow{5}{*}{GT-\{GPT4-V\}} & w/o. semantic patch & 35.24 & 18.48 & 9.32  & 5.07 & 2.80 & 18.01 & 19.83 & 20.21 & 42.99 \\ 
& w/. four equal           & 37.03 & 19.54 & 9.90  & 5.33 & 3.42 & 18.51 & 20.13 & 20.77 & 43.27 \\ 
& w/o. semantic filtering  & 36.72 & 19.60 & 10.21 & 5.77 & 3.56 & 18.39 & 20.02 & 20.13 & 43.08 \\ 
& w/o. hierarchical aggregation     & 13.61 & 7.18  & 3.74  & 2.07 & 0.62 & 11.79 & 16.98 & 18.25 & 40.37 \\ 
& Ours               & \textbf{37.30} & \textbf{20.07} & \textbf{10.62} & \textbf{6.00} & \textbf{3.60} & \textbf{18.67} & \textbf{20.39} & \textbf{20.99} & \textbf{43.33} \\ 
   \bottomrule
\end{tabular}
\caption{Ablation study of each component on DID-Bench, GT-\{LLaVA\} and GT-\{GPT4-V\} refer to ground truth captions generated by LLaVA and GPT-4Vision, respectively.}
\label{tab:table8}
\end{table*}

\noindent \textbf{Cost Analysis.}
Our method is completely training-free with lower cost than expected. Despite multiple MLLM and LLM calls, our strategy leverages small patch sizes  to reduce tokens, while  batch processing and vLLM\cite{kwon2023vllm} accelerates inference, ensuring efficiency. Compared to training-based MLLMs and other training-free methods, our cost remains acceptable (\cref{tab:cost}).  
\begin{table}[b]
   \centering
    \small
    \vspace{-3mm}
     \setlength{\belowcaptionskip}{-3mm}
        \renewcommand{\arraystretch}{0.6}
    
      \setlength{\tabcolsep}{2pt}
    \begin{tabular}{c|c|c}
        \toprule
         Method & Training& Inference \\
         \midrule   
LLaVA1.6  &   Hundreds of A100 GPU Hours & 0.55s per image\\
IT     & None &  2.80s per image\\
Ours  & None &  7.04s per image \\
   \bottomrule
\end{tabular}
\caption{Cost analysis of inference per image using 8$\times$A40.}
\label{tab:cost}
\end{table}

\subsection{Ablation study}

To evaluate each component’s significance and effectiveness, we conducted an ablation study on the DID-Bench, as shown in \cref{tab:table8}. Removing the semantic patch description dropped performance across metrics, highlighting the semantic patch description is crucial for capturing essential details of core elements within an image. Additionally, dividing the image into four equal patches led to a decrease in performance, likely due to incomplete division and an increased likelihood of incorrect descriptions. Removing the semantic filtering method also lowered descriptive quality, indicating that this method improves the reliability and accuracy of the descriptions. Finally, replacing hierarchical aggregation with a direct fusion approach like PoCa \cite{chen2024poca}, resulted in a significant drop, possibly due to the simultaneous aggregation of data from multiple patches, leading to redundant information and insufficient fusion.

%% file: 5_Conclution.tex
\section{Conclusion}

We propose a \textit{divide-then-aggregate} approach, a training-free pipeline that enhances image captioning by capturing more details and reducing hallucinations. This method divides images into semantic and spatial components, generating candidate captions to improve local perception. It then applies a hierarchical aggregation strategy, combining intra-patch and inter-patch aggregation with semantic filtering to enhance the global quality and accuracy of the captions. Extensive experiments on both open-source and closed-source models show our method achieves competitive performance compared to fine-tuned models. In the future, we aim to reduce the computational cost of this method, as it requires multiple descriptions and LLM computations to generate and integrate a complete caption. Addressing this issue is crucial for enabling the rapid generation of large-scale image-caption pairs.

\section{Acknowledge}
This work is sponsored by CCF-Zhipu.AI Large Model Innovation Fund.

%% file: X_suppl.tex
\clearpage
\setcounter{page}{1}
\maketitlesupplementary
\setcounter{section}{0}
\renewcommand\thesection{\Alph{section}}
\setlength{\cftbeforesecskip}{1.2em} % 一级标题之间的行距
\setlength{\cftbeforesubsecskip}{0.3em} % 二级标题之间的行距
% \renewcommand{\cftsecleader}{\cftdotfill{\cftdotsep}} % 设置目录中的点线样式

% \begin{document}

% \tableofcontents
% \section{Rationale}
% \label{sec:rationale}
% % 
% Having the supplementary compiled together with the main paper means that:
% % 
% \begin{itemize}
% \item The supplementary can back-reference sections of the main paper, for example, we can refer to \cref{sec:intro};
% \item The main paper can forward reference sub-sections within the supplementary explicitly (e.g. referring to a particular experiment); 
% \item When submitted to arXiv, the supplementary will already included at the end of the paper.
% \end{itemize}
% % 
% To split the supplementary pages from the main paper, you can use \href{https://support.apple.com/en-ca/guide/preview/prvw11793/mac#:~:text=Delete%20a%20page%20from%20a,or%20choose%20Edit%20%3E%20Delete).}{Preview (on macOS)}, \href{https://www.adobe.com/acrobat/how-to/delete-pages-from-pdf.html#:~:text=Choose%20%E2%80%9CTools%E2%80%9D%20%3E%20%E2%80%9COrganize,or%20pages%20from%20the%20file.}{Adobe Acrobat} (on all OSs), as well as \href{https://superuser.com/questions/517986/is-it-possible-to-delete-some-pages-of-a-pdf-document}{command line tools}.
\section{Experiments Details}

\subsection{Metrics Details}

\textbf{BLEU.} The BLEU \cite{papineni2002bleu} score was introduced for machine translation. This metric measures the similarity between model-generated text and reference text. BLEU score typically calculates consistent n-grams between the two texts. A higher score indicates greater similarity between the texts.

\noindent \textbf{ROUGE.} The ROUGE \cite{lin2004rouge} score was initially designed to evaluate summarization. ROUGE measures the overlap of n-grams. Its variant, ROUGE-L, calculates the similarity between generated text and reference text based on the longest common subsequence.

\noindent \textbf{METEOR.} METEOR \cite{lavie-agarwal-2007-meteor} is another machine translation metric. It is based on the harmonic mean of unigram precision and recall between generated sentences and reference sentences, with improved comparison in aspects such as synonyms, and stem variations.

\noindent \textbf{CIDEr.} CIDEr \cite{vedantam2015cider} evaluates image descriptions using TF-IDF (Term Frequency-Inverse Document Frequency) weighted n-grams, calculating cosine similarity between candidate captions and reference captions, incorporating both precision and recall.

\noindent \textbf{SPICE.} SPICE \cite{anderson2016spice} is a semantic-based metric for image description evaluation. It converts image descriptions into scene graphs, extracting objects, attributes, and their relationships to measure semantic similarity between candidate and reference descriptions.

\noindent \textbf{WMD.} WMD \cite{DBLP:conf/eacl/KilickayaEIE17} is a semantic text similarity metric that uses word vectors to represent words and computes a distance measure to determine the similarity between two text sequences.

\noindent \textbf{CAPTURE.} CAPTURE \cite{dong2024capture} evaluates image descriptions by identifying key visual elements. It uses a scene graph parser to extract objects, attributes, and relationships from both candidate and reference texts. Abstract nouns are filtered using a stop-word list. F1 scores are calculated based on exact matches, synonym matches, and soft matches, aligning with human evaluation standards. The final CAPTURE score is a weighted combination of these F1 scores, defined as:
\begin{equation}
\text{CAPTURE} = \frac{\alpha \cdot F1_{obj} + \beta \cdot F1_{attr} + \gamma \cdot F1_{rel}}{\alpha + \beta + \gamma},
\end{equation}
where $\alpha = 5, \ \beta = 5, \ \gamma = 2.$

\noindent \textbf{Polos.} Polos\cite{wada2024polos} is a supervised automatic evaluation metric, which computes scores from multimodal inputs using a parallel feature extraction mechanism that leverages embeddings trained through large-scale contrastive learning. It is designed to better align with human judgments and handle diverse images and texts.

\noindent \textbf{CHAIR.} CHAIR\cite{rohrbach2018chair_metric} is designed for evaluating object hallucination in image captions. It calculates the proportion of objects that appear in a caption but not in an image. Its two variants, CHAIRi and CHAIRs, evaluate the hallucination at the object instance level and the sentence level, respectively. They are calculated as follows:
\begin{equation}
\text{CHAIRi} = \frac{\{ \text{hallucinated objects} \}}{\{ \text{all objects mentioned} \}},
\end{equation}

\begin{equation}
\text{CHAIRs} = \frac{\{ \text{sentences with hallucinated objects} \}}{\{ \text{all sentences} \}}.
\end{equation}

\noindent \textbf{LIN-Bench.} LIN-Bench \cite{pi2024image-textual} is an evaluation framework introduced in the article to assess the readability and linguistic complexity of generated image descriptions. It uses metrics such as ARI, FK, and SMOG. ARI focuses on the number of words in a sentence and the average number of characters per word, FK is based on sentence length and syllable count, and SMOG measures the use of polysyllabic words. Higher scores on these metrics typically indicate that the text contains more information and detail.

\noindent \textbf{CLIP-score and DINO-score.} CLIP-score utilizes the CLIP \cite{radford2021Clip} model to extract image embeddings and calculate cosine similarity between the generated image and a candidate image. DINO-score uses the DINOv2 \cite{oquab2023dinov2} model to extract features from both images and compute cosine similarity. CLIP is trained on image-text datasets and captures semantic features of images, allowing CLIP-score to reflect high-level semantic similarity. DINOv2 is a self-supervised vision model that effectively captures fine-grained visual features, making DINO-score well-suited for assessing detailed visual similarity.

\noindent \textbf{VQA.} The VQA task \cite{chen2024poca} is designed to assess caption quality in conveying image content. We conducted this experiment using 625 images from the VQA-V2 validation set \cite{goyal2017vqa-v2} (5,000 questions total). A text-only LLM \cite{dubey2024llama} answered questions based on captions from various methods.

\noindent \textbf{POPE.} POPE \cite{li2023pope} is a mainstream evaluation metric for multimodal models, primarily focusing on object-level hallucinations. It employs three polling strategies: sampling objects randomly, selecting from popular objects, and choosing among frequently co-occurring objects that do not exist in the image, which is referred to as adversarial sampling. We conduct our evaluation on the MSCOCO \cite{lin2014COCO-data} validation dataset, which consists of 500 images, each accompanied by 6 questions. The evaluation metrics include Accuracy, Precision, Recall, and F1 score.
\subsection{Prompt for Semantic Filtering}
\begin{figure*}[h]
  \centering
   \includegraphics[width=1\linewidth]{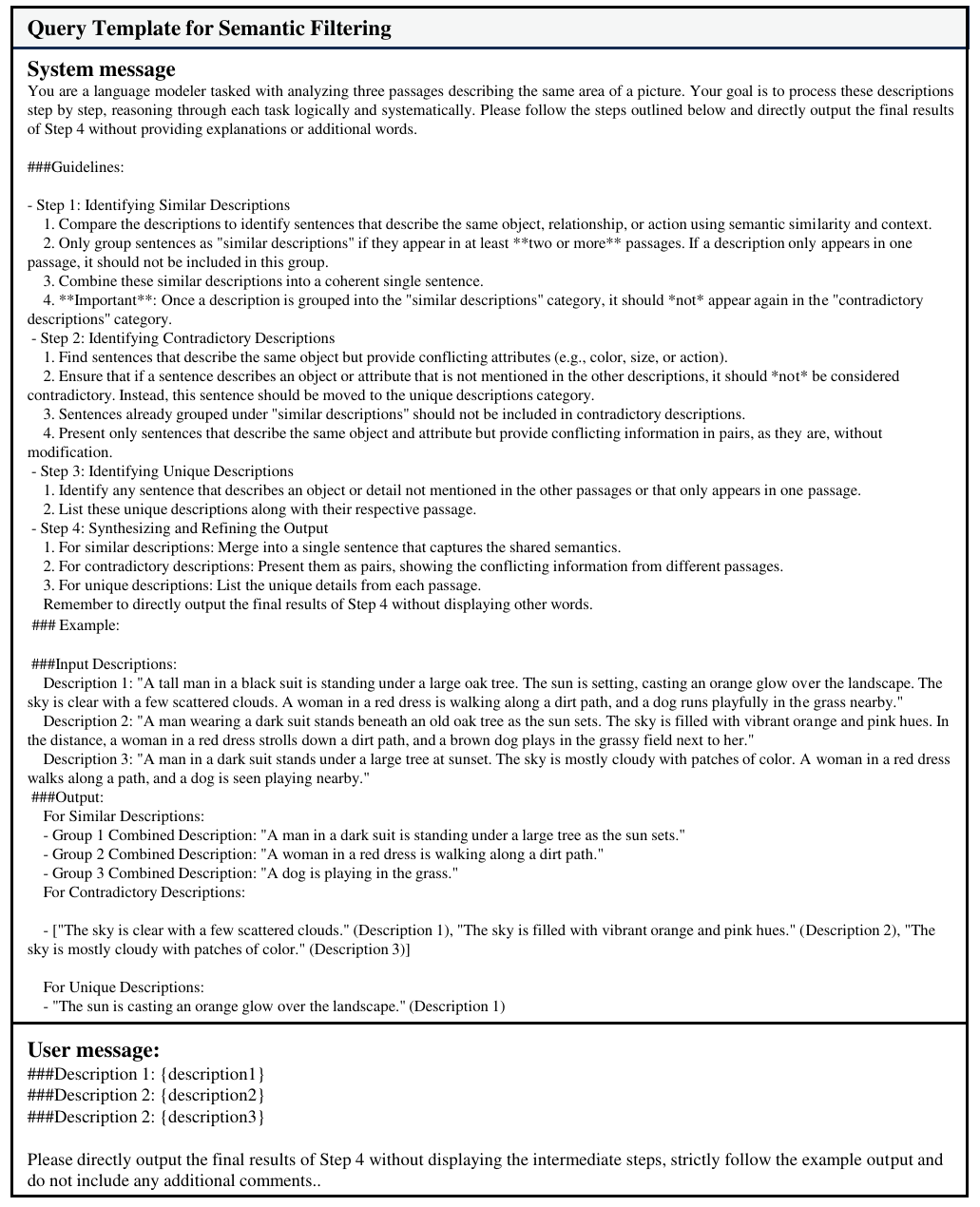}
    
   \caption{The system and user prompts used for  semantic filtering query.}
   \label{fig:prompt1}
\end{figure*}

In \cref{fig:prompt1}, we demonstrate the detailed prompt used to guide the LLM in analyzing the semantic content of candidate descriptions and categorizing them into three groups. First, we specify the task goal for the LLM. Then, we emphasize key points to remember during the extraction process: 1) Identify sentences that describe the same object across different descriptions and consolidate them into a single sentence; 2) Identify contradictory sentences from different descriptions; 3) Identify sentences that appear only in one description but describe important objects. We also emphasize that each sentence should belong to only one category. Finally, we provide the LLM with manually labeled contextual examples to enhance its ability to follow instructions.
\begin{table}[b]
    \centering
    \small
    % \vspace{-2.5em}
        % \setlength{\abovecaptionskip}{0mm}
     \setlength{\belowcaptionskip}{-2mm}
        \renewcommand{\arraystretch}{0.8}
      \setlength{\tabcolsep}{1pt}
    \begin{tabular}{cc|ccccc}
      \toprule
         IoU& Blip2score &  CIDEr & METEOR & ROUGE & SPICE & WMD \\
        % \hline

\midrule

0.3&0.2 & 4.03 &	19.16 	&20.76 &	20.89 &	44.40  \\ 
0.3&0.3 &3.51 &	19.36 &	20.77 &	21.09 &	44.31  \\ 
0.3&0.4 &  4.18 &	18.54 &	20.52 &	20.97& 	44.21 \\
\midrule
% \hline
0.4&0.2 & 3.35 &	19.85 &	21.00 &	21.16 &	44.49  \\ 
0.4&0.3 & 4.55 &	19.69 &	21.04 &	21.39 	&44.64 \\ 
0.4&0.4 & 4.12 	&19.21 	&20.83 	&21.10 &	44.46  \\
% \hline
% \midrule
%  \multicolumn{2}{c|}{LLaVA-1.6}&     0.74& 	14.18 &	19.86 &	19.79 &	42.24 \\
% % \midrule
% % \hline
% % 0.5&0.2 & 3.12 &	\textbf{19.87} 	&20.64 &	20.88 	&44.49 \\ 
% % 0.5&0.3 & \textbf{4.58} 	&19.70 &	\textbf{21.15} &	21.29 &	44.61 \\
% % 0.5&0.4 & 3.43 	&19.46 &	20.82 	&21.14 &	44.55  \\ 
% \midrule
% % \hline
% \multicolumn{2}{c|}{w/. 2 spatial patches} & 1.97&15.47&18.98&19.74&42.80\\
% \multicolumn{2}{c|}{w/. object level} & 3.11&20.67&20.88&21.40&45.04\\
% \multicolumn{2}{c|}{w/. object level} & 3.11&20.67&20.88&21.40&45.04\\
% 
   \bottomrule
\end{tabular}
\caption{Ablation study of Blip2Score and IoU  on DID-Bench.}
\label{tab:iou}
\end{table}
\begin{table}[b]
    \centering
    \small
    % \vspace{-2.5em}
        % \setlength{\abovecaptionskip}{0mm}
     \setlength{\belowcaptionskip}{-2mm}
        \renewcommand{\arraystretch}{0.8}
      \setlength{\tabcolsep}{3pt}
    \begin{tabular}{c|ccccc}
      \toprule
     
         Method &  CIDEr & METEOR & ROUGE & SPICE & WMD \\
        % \hline
        \midrule
 \rowcolor{gray!20}
        \multicolumn{6}{c}{\textit{\textbf{LLaVA-1.6}}} \\
Global&     0.74& 	14.18 &	19.86 &	19.79 &	42.24 \\
% \midrule
% \hline
% 0.5&0.2 & 3.12 &	\textbf{19.87} 	&20.64 &	20.88 	&44.49 \\ 
% 0.5&0.3 & \textbf{4.58} 	&19.70 &	\textbf{21.15} &	21.29 &	44.61 \\
% 0.5&0.4 & 3.43 	&19.46 &	20.82 	&21.14 &	44.55  \\ 
% \midrule
% \hline
w/. 2 spatial patches & 1.97&15.47&18.98&19.74&42.80\\
w/. object level & 3.11&20.67&20.88&21.40&45.04\\
ours & 4.55&19.69&21.04&21.39&44.64\\
   \bottomrule
\end{tabular}
\caption{Ablation study of patch numbers and objec level description on DID-Bench.}
\label{tab:patch number}
\end{table}

\begin{table}[b]
    \centering
    \small
    % \vspace{-2.5em}
        % \setlength{\abovecaptionskip}{0mm}
     \setlength{\belowcaptionskip}{-5mm}
        \renewcommand{\arraystretch}{0.8}
      \setlength{\tabcolsep}{3pt}
    \begin{tabular}{c|ccccc}
      \toprule
     
         Method &  CIDEr & METEOR & ROUGE & SPICE & WMD \\
        % \hline
        \midrule
 \rowcolor{gray!20}
        \multicolumn{6}{c}{\textit{\textbf{Cambrian}}} \\

% \midrule
% \hline
% 0.5&0.2 & 3.12 &	\textbf{19.87} 	&20.64 &	20.88 	&44.49 \\ 
% 0.5&0.3 & \textbf{4.58} 	&19.70 &	\textbf{21.15} &	21.29 &	44.61 \\
% 0.5&0.4 & 3.43 	&19.46 &	20.82 	&21.14 &	44.55  \\ 
% \hline
% 
{Global} & 0.00 	&6.77 &	12.62 &	10.31 	&35.06 \\ 
{Ours+Cambrian}                & 3.31 & 15.47 & 18.41 & 16.27 & 39.95 \\
% \hline

 \rowcolor{gray!20}
        \multicolumn{6}{c}{\textit{\textbf{CogVLM}}} \\
{Global} & 0.00 &	7.89 &	13.78 	&15.78& 	38.65  \\ 
{Ours+CogVLM}       & 1.25 & 14.49 & 18.00 & 18.90 & 41.88 \\ 
   \bottomrule
\end{tabular}
\caption{Visually Enhanced MLLMs on DID-Bench.}
\label{tab:Visually Enhanced MLLMs}
\end{table}
\subsection{Prompt for Aggregation}
\begin{figure*}[h]
  \centering
   \includegraphics[width=1\linewidth]{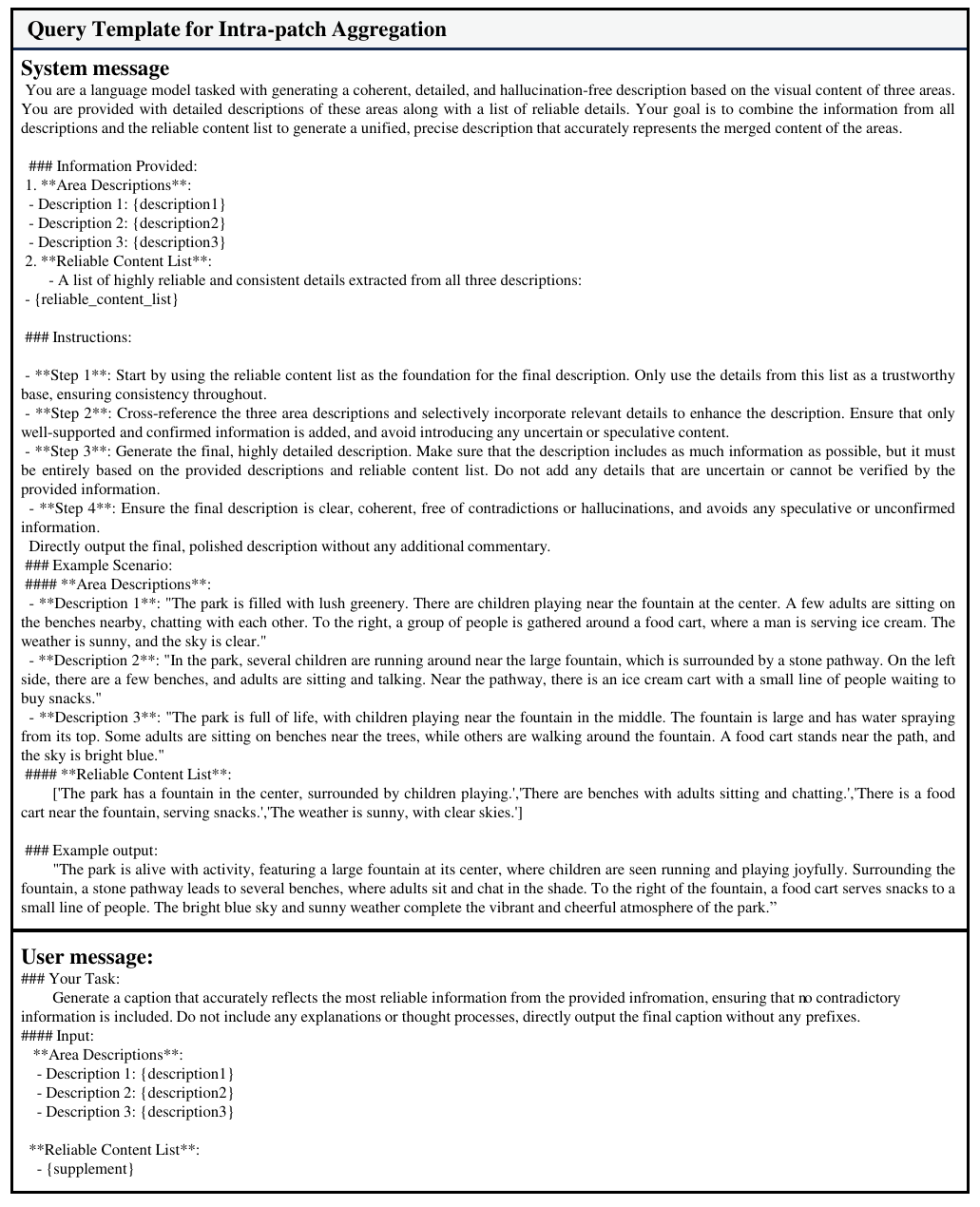}
    
   \caption{The system and user prompts used for  intra-patch aggregation query.}
   \label{fig:prompt2}
\end{figure*}

\begin{figure*}[h]
  \centering
   \includegraphics[width=1\linewidth]{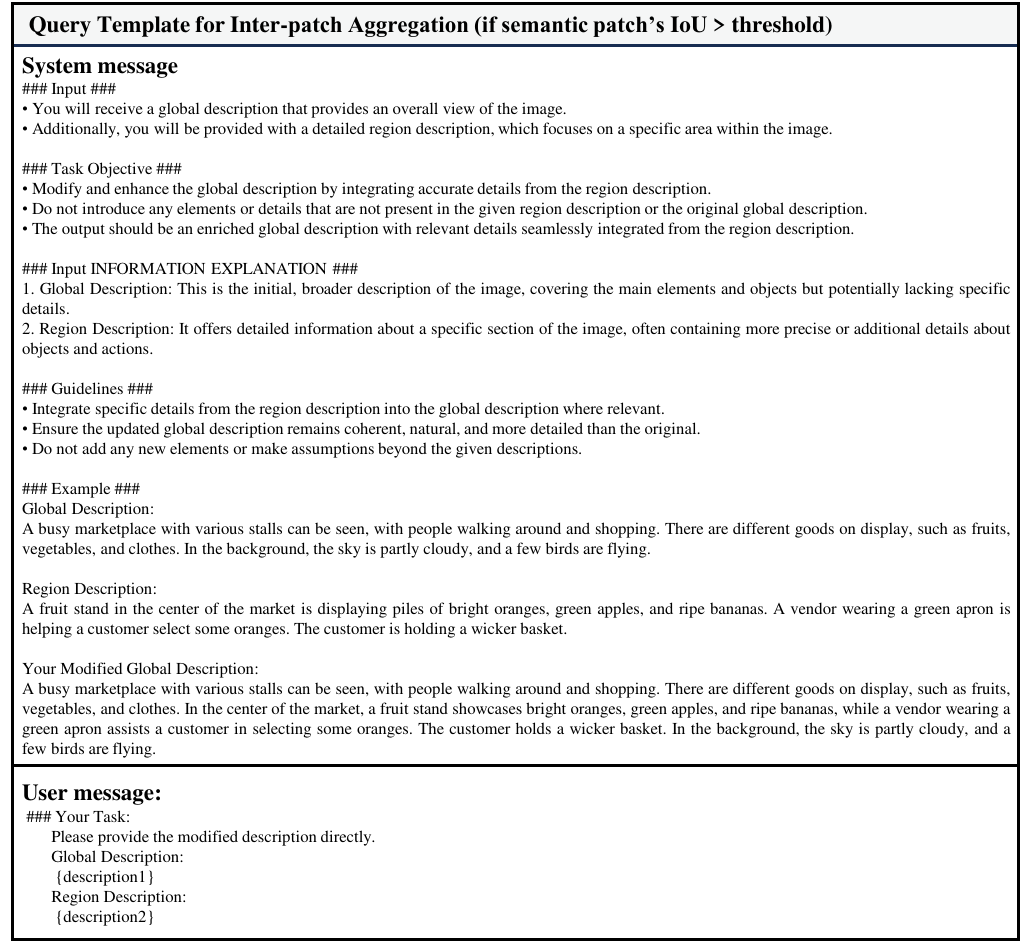}
    
   \caption{The system and user prompts used for  inter-patch aggregation (if semantic patch's IoU \textgreater 
  threshold) query.}
   \label{fig:prompt6}
\end{figure*}

\begin{figure*}[h]
  \centering
   \includegraphics[width=1\linewidth]{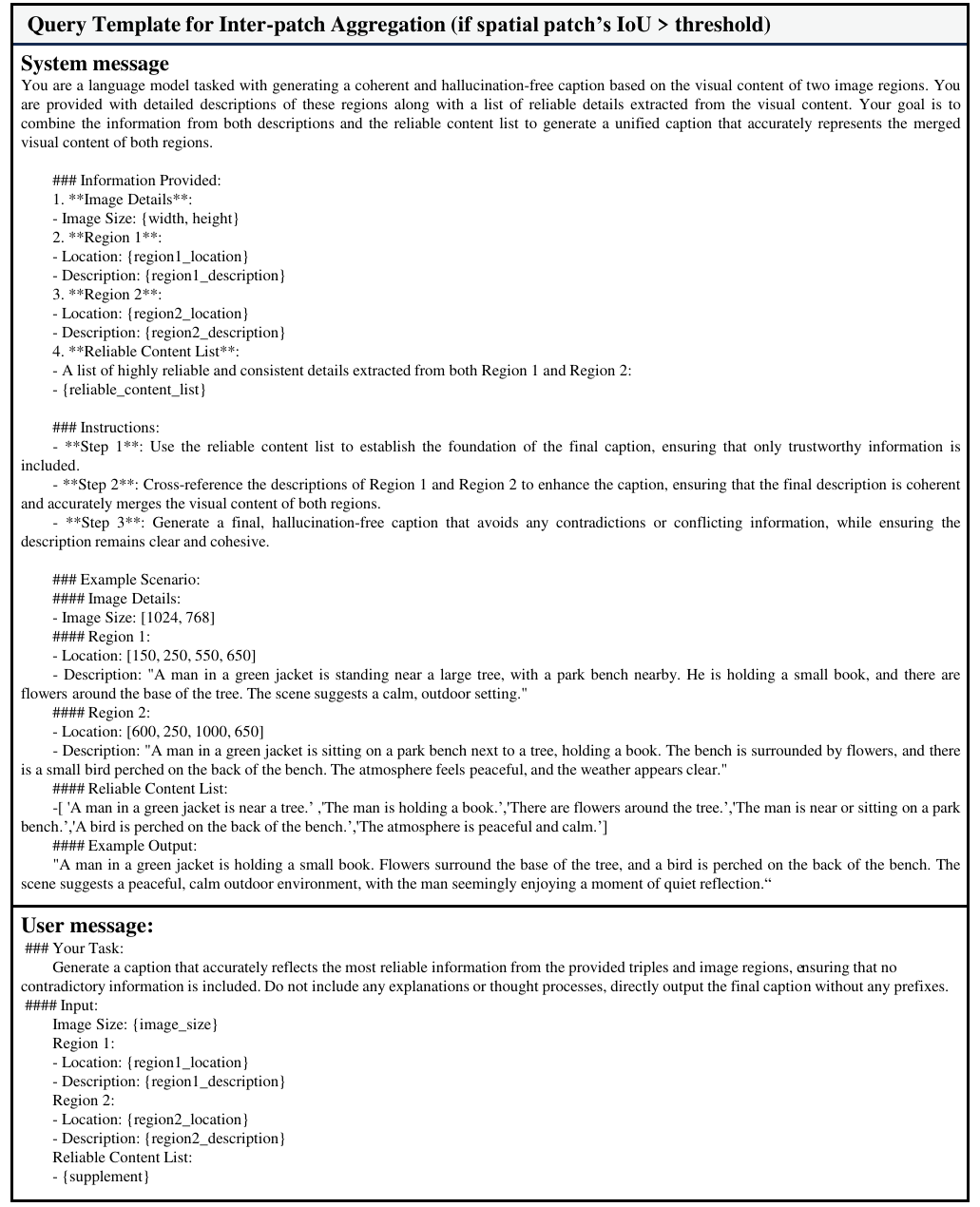}
    
   \caption{The system and user prompts used for  inter-patch aggregation (if spatial patch's IoU \textgreater 
  threshold) query.}
   \label{fig:prompt3}
\end{figure*}

\begin{figure*}[h]
  \centering
   \includegraphics[width=1\linewidth]{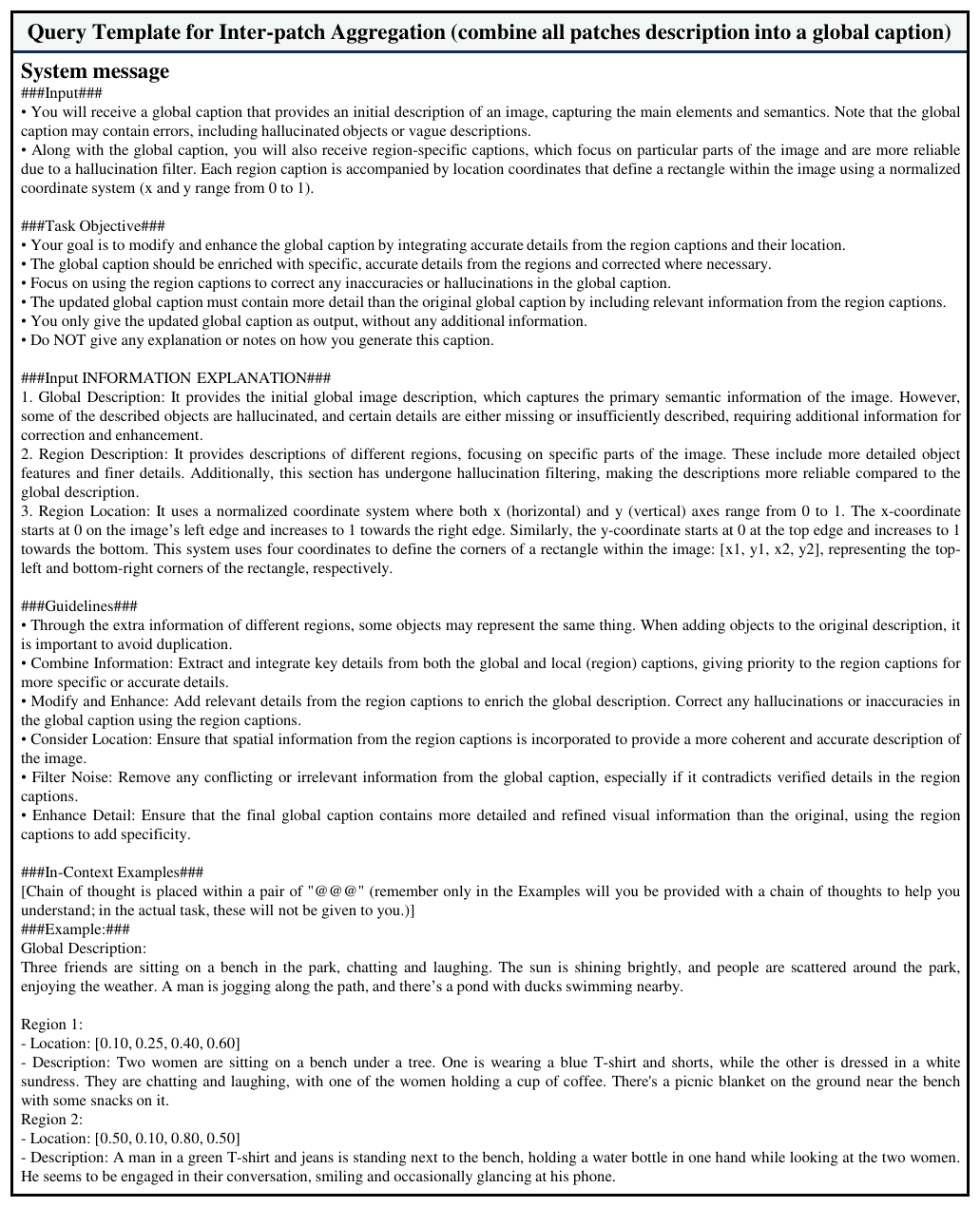}
    
   \caption{The system prompts used for  inter-patch aggregation (combine all patches description into a global caption) query (first half).}
   \label{fig:prompt4}
\end{figure*}

\begin{figure*}[t]
  \centering
   \vspace{80pt}
   \includegraphics[width=1\linewidth]{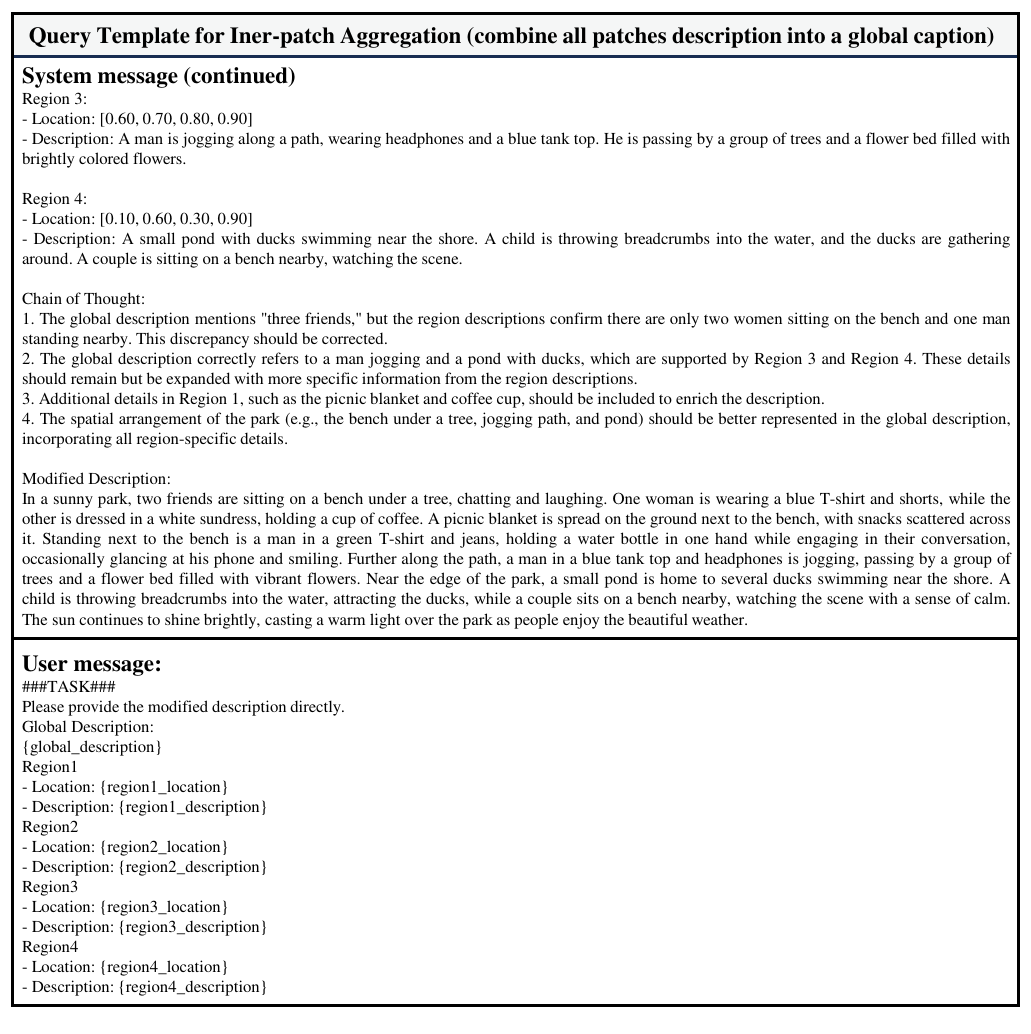}
    
   \caption{The system and user prompts used for  inter-patch aggregation (combine all patches description into a global caption) query (second half).}
   \label{fig:prompt5}
\end{figure*}

% 我们在 \cref{fig:prompt2,fig:prompt3,fig:prompt4,fig:prompt5}中展示了用于aggregation的LLM prompt。其中\cref{fig:prompt2}是intra-Patch aggregation，需要将同一个Patch描述合并成一个描述，在这个例子中，我们输入三个候选描述以及由semantic filtering得到的可信度较高的描述，并指示llm根据可信度较高的描述合并成一个描述，确保描述的准确性以及避免重复。\cref{fig:prompt3,fig:prompt4}是当spatial patches中有iou大于某个阈值时进行的融合prompt，他会将两个region的描述根据semantic filtering得到的可信描述列表得到这两个区域合并后的整体描述。最后\cref{fig:prompt5}举了一个如何将不同spatial Patch的描述合并成一个Global描述的例子。在这里我们是假设四个spatial Patch相互之间都没有超过iou阈值，所以我们有四个Patch描述和一个global描述作为输入，我们提示llm根据可信度较高的Patch描述进行信息的补充和纠正global描述中潜在的幻觉。
In \cref{fig:prompt2,fig:prompt6,fig:prompt3,fig:prompt4,fig:prompt5}, we demonstrate LLM prompts used for aggregation. \cref{fig:prompt2} shows intra-patch aggregation, where descriptions of the same patch are merged into a single description. In this example, we input three candidate descriptions along with a high-confidence description obtained through semantic filtering and instruct the LLM to merge them based on the high-confidence description, ensuring accuracy and avoiding redundancy. \cref{fig:prompt6} show prompts for merging when the IoU between the semantic patch and the global image exceeds a certain threshold. The LLM  combines descriptions of the key semantic regions based on the high-confidence descriptions obtained through semantic filtering, filling in the missing parts of the global image and correcting any errors within them. \cref{fig:prompt3} and \cref{fig:prompt4} show prompts for merging when the IoU between spatial patches exceeds a certain threshold. The LLM combines the descriptions of the two regions based on the high-confidence descriptions obtained through semantic filtering, resulting in a unified description for the merged region. Finally, \cref{fig:prompt5} provides an example of how to merge descriptions from different spatial patches into a global description. Here, we assume that the IoU between the four spatial patches is below the threshold, so we have four patch descriptions and one global description as input. We prompt the LLM to use the high-confidence patch descriptions to supplement and correct potential hallucinations in the global description.

\section{Additional Results and Experiments}
\subsection{Ablation Study in DID-Bench}
To evaluate our method regarding the selection of Blip2Score and IoU thresholds, we employ grid search for the experiments, as shown in the \cref{tab:iou}. The small variation in the metrics indicates that the nearby thresholds are not sensitive to the experimental results and all surpass the vanilla MLLM. In addition, we have carried out experiments on two aspects: one is solely splitting the image into two patches, and the other is adding object-level information. The results presented in \cref{tab:patch number} demonstrate that dividing the image into two patches hinders the model from acquiring perception, while adding object-level information extracted by GRiT\cite{wu2022grit} is beneficial. 
\subsection{Experiment on Visually Enhanced MLLMs}
To assess the effectiveness of our method on MLLMs with stronger visual capabilities, we conducted experiments on the CogVLM\cite{wang2025cogvlm} and Cambrian\cite{tong2025cambrian}. The results are presented in \cref{tab:Visually Enhanced MLLMs}. As indicated by the data in the table, our method continues to significantly enhance performance on these models.
\subsection{CAPTURE Score in DID-Bench}
We also evaluated the CAPTURE score on DID-Bench  \cite{pi2024image-textual}, as shown in \cref{tab:table9} and \cref{tab:table10}. We present the CAPTURE scores for both open-source and closed-source large models, and it is clear that our method consistently improves the performance of these models on this metric. An interesting observation is that LLaVA and Mini-Gemini outperform many closed-source large models in the CAPTURE score, which could be attributed to their use of the GPT-4V-annotated ShareGPT4V \cite{chen2023sharegpt4v} dataset during training. This further highlights the critical role of high-quality image-text descriptions in improving model performance.
% 我们还在DID-Bench上评测了CAPTURE score，如表7和表8所示。我们展示了开源大模型和闭源大模型CAPTURE分数的表现，并可以清楚地看到我们的方法依然普遍提升了这些模型在该分数上的表现的提高。一个很有趣的现象是，LLaVA和Minigemini本身的表现超过了众多闭源大模型，这可能是因为他们使用了GPT-4V标注的数据ShareGpt4V用于训练，这进一步说明了高质量的图文描述对于提升模型性能的重要性。

\subsection{Experiment on LIN-Bench}

We evaluated our method on LIN-Bench \cite{pi2024image-textual} using images from DID-Bench. LIN-Bench focuses on readability and descriptive detail to assess the quality and complexity of generated text. We also conducted a statistical analysis of the description lengths generated by different methods on the DID-Bench dataset, as shown in the \cref{tab:count}. Since this benchmark is only suitable for descriptions longer than 100 words, we did not use PoCa \cite{chen2024poca} as a baseline. As shown in \cref{tab:table11}, our method achieves higher scores across LIN-Bench metrics (ARI, FK, SMOG), demonstrating that it produces detailed descriptions.

\begin{table}[ht]
    \centering
       \renewcommand{\arraystretch}{0.86}

     \setlength{\tabcolsep}{3pt}
    \small
    \begin{tabular}{l |cccc}
        \toprule
         Description & CAPTURE & $F1_{obj}$ & $F1_{attr}$ & $F1_{rel}$ \\
        \midrule
         LLaVA1.5 &49.81 &	59.17 &	38.26 &	55.25    \\
         LLaVA1.5+IT \cite{pi2024image-textual} &53.86 &	61.96 	&44.74 &	56.67    \\
         LLaVA1.5+PoCa \cite{chen2024poca} &44.84 &	56.21 &	31.27& 	50.37  \\
         LLaVA1.5+Syn \cite{dong2024capture} & 58.51 &	65.74 &	52.24 &	56.12  \\
         LLaVA1.5+Ours & \textbf{59.61} & \textbf{66.78} & \textbf{53.11} & \textbf{57.95}  \\
        \midrule
         LLaVA1.6 & 59.58 	&65.66& 	54.12& 	58.01    \\
         LLaVA1.6+IT \cite{pi2024image-textual} & 61.60 &	67.56& 	56.68 &	59.00    \\
         LLaVA1.6+PoCa \cite{chen2024poca} &54.53 &	60.54 &	49.11 &	53.02   \\
         LLaVA1.6+Syn \cite{dong2024capture} & 62.67 &	67.93 &	59.17 &	58.31   \\
         LLaVA1.6+Ours & \textbf{64.48 } & \textbf{69.61} & \textbf{61.27} & \textbf{59.71}  \\
         \midrule
         Mini-Gemini & 62.28 &	67.46 	&59.00& 	57.58    \\
         Mini-Gemini+IT \cite{pi2024image-textual} &63.24 &	68.13 &	60.36 &	58.16   \\
         Mini-Gemini+PoCa \cite{chen2024poca} &55.05 &	60.54 &	50.65 &	52.49  \\
         Mini-Gemini+Syn \cite{dong2024capture} & 64.22 &	68.75 &	61.84 &	58.85   \\
         Mini-Gemini+Ours & \textbf{65.55} & \textbf{69.69} & \textbf{63.75} & \textbf{59.86}  \\
        \bottomrule
    \end{tabular}
    \normalsize 
    \caption{CAPTURE scores of open-source models on DID-
Bench are
weighted combinations of various F1 scores, where higher F1
scores for objects, attributes, and relations indicate better perfor-
mance in capturing these aspects.}
    \label{tab:table9}
\end{table}
\begin{table}[t]
    \centering
       \renewcommand{\arraystretch}{0.86}

     \setlength{\tabcolsep}{3pt}
    \small
    \begin{tabular}{l |cccc}
        \toprule
         Description & CAPTURE & $F1_{obj}$ & $F1_{attr}$ & $F1_{rel}$ \\
        \midrule
         GLM-4V-Plus &56.04 &	64.12 &	47.84 &	56.30    \\
         GLM-4V-Plus+IT \cite{pi2024image-textual} &59.37 &	66.75 	&52.50 &	58.10   \\
         GLM-4V-Plus+PoCa \cite{chen2024poca} &53.25 &	59.04& 	47.60 &	52.88  \\
         GLM-4V-Plus+Syn \cite{dong2024capture} & 62.91 &	68.57 	&59.17 	&58.09  \\
         GLM-4V-Plus+Ours & \textbf{64.72} & \textbf{70.55} & \textbf{60.83} & \textbf{59.89}  \\
        \midrule
         GPT-4o & 57.40& 	64.39& 	50.46 &	57.29    \\
         GPT-4o+IT \cite{pi2024image-textual} & 60.53 &	67.26& 	54.48 &	58.83    \\
         GPT-4o+PoCa \cite{chen2024poca} &50.40 &	56.87& 	43.81 &	50.70    \\
         GPT-4o+Syn \cite{dong2024capture} & 62.93& 	68.98 	&58.89 	&57.93   \\
         GPT-4o+Ours & \textbf{63.28 } & \textbf{68.97} & \textbf{59.07} & \textbf{59.57}  \\
         \midrule
         Claude-3.5 & 51.76 &	52.14 &	53.43 &	46.65    \\
         Claude-3.5+IT \cite{pi2024image-textual} &64.23 &	68.68 &	62.29 &	57.98    \\
         Claude-3.5+PoCa \cite{chen2024poca} &55.83 &	59.92 	&53.32 	&51.86   \\
         Claude-3.5+Syn \cite{dong2024capture} & 63.66 &	67.92 	&61.94& 	57.30   \\
         Claude-3.5+Ours & \textbf{64.72} & \textbf{69.21} & \textbf{62.77} & \textbf{58.37}  \\
        \bottomrule
    \end{tabular}
    \normalsize 
    \caption{CAPTURE scores of close-source models on DID-
Bench are
weighted combinations of various F1 scores, where higher F1
scores for objects, attributes, and relations indicate better perfor-
mance in capturing these aspects.}
    \label{tab:table10}
\end{table}
\begin{table}[ht]
    \centering
    \small
    \begin{tabular}{l |cccc}
        \toprule
        Description & ARI & FK & SMOG & Avg \\
        \midrule
         LLaVA1.5 &8.69 &	8.18 &	10.80 &	9.22   \\
         LLaVA1.5+IT \cite{pi2024image-textual}&8.85 &	8.35 &	10.86 &	9.35   \\
         LLaVA1.5+Syn \cite{dong2024capture}&9.74 &	8.93 &	10.97 &	9.88    \\
         LLaVA1.5+Ours & \textbf{11.50} 	&\textbf{10.42} 	&\textbf{12.24}&\textbf{11.38 }   \\
        \midrule
         LLaVA1.6 &9.71 &	9.19 &	11.41& 	10.10   \\
         LLaVA1.6+IT \cite{pi2024image-textual}&10.05 	&9.48 &	11.58 &	10.37   \\
         LLaVA1.6+Syn \cite{dong2024capture}& 10.75 &	9.85 &	11.79 &	10.80    \\
         LLaVA1.6+Ours & \textbf{12.49} &	\textbf{11.37} 	&\textbf{12.98}	&\textbf{12.28}  \\
         \midrule
         Mini-Gemini &9.31 	&8.55 &	10.87 &	9.57  \\
         Mini-Gemini+IT \cite{pi2024image-textual}& 9.52 &	8.75 &	10.97 &	9.74   \\
         Mini-Gemini+Syn \cite{dong2024capture}& 10.69 &	9.57 &	11.35 &	10.53  \\
         Mini-Gemini+Ours & \textbf{12.11} &	\textbf{10.77} &	\textbf{12.31} &	\textbf{11.73}  \\
        \bottomrule
    \end{tabular}
    \normalsize 
    \caption{LIN-Bench Results. Our outputs contain a higher number of syllables and characters.}
    \label{tab:table11}
\end{table}
% 1211.485	12.29	245.815
% 587.205	7.265	128.08
% 709.77	8.05	155.475
% 723.615	7.725	156.12
% 275.22	2.525	59.725
% 1044.4	9.525	221.695
\begin{table}[ht]
    \centering
    \small
    \begin{tabular}{l |ccc}
        \toprule
        Description &Chars&	Sentences &Words	 \\
        \midrule
        Ground\_truth &1211.49	&12.29&	245.82\\
         LLaVA1.6 &587.21	&7.27	&128.08   \\
         LLaVA1.6+IT \cite{pi2024image-textual}&709.77&	8.05	&155.48   \\
         LLaVA1.6+Syn \cite{dong2024capture}&723.62&7.73	&156.12   \\
          LLaVA1.6+PoCa \cite{chen2024poca}&275.22&	2.53&	59.73    \\
         LLaVA1.6+Ours & 1044.40&	9.53	&221.70  \\
        \bottomrule
    \end{tabular}
    \normalsize 
    \caption{Statistical comparison of image description lengths generated by different methods on the DID-bench dataset, measured in average characters, words, and sentences per description.}
    \label{tab:count}
\end{table}
\begin{figure}[h]
  \centering
   \includegraphics[width=1\linewidth]{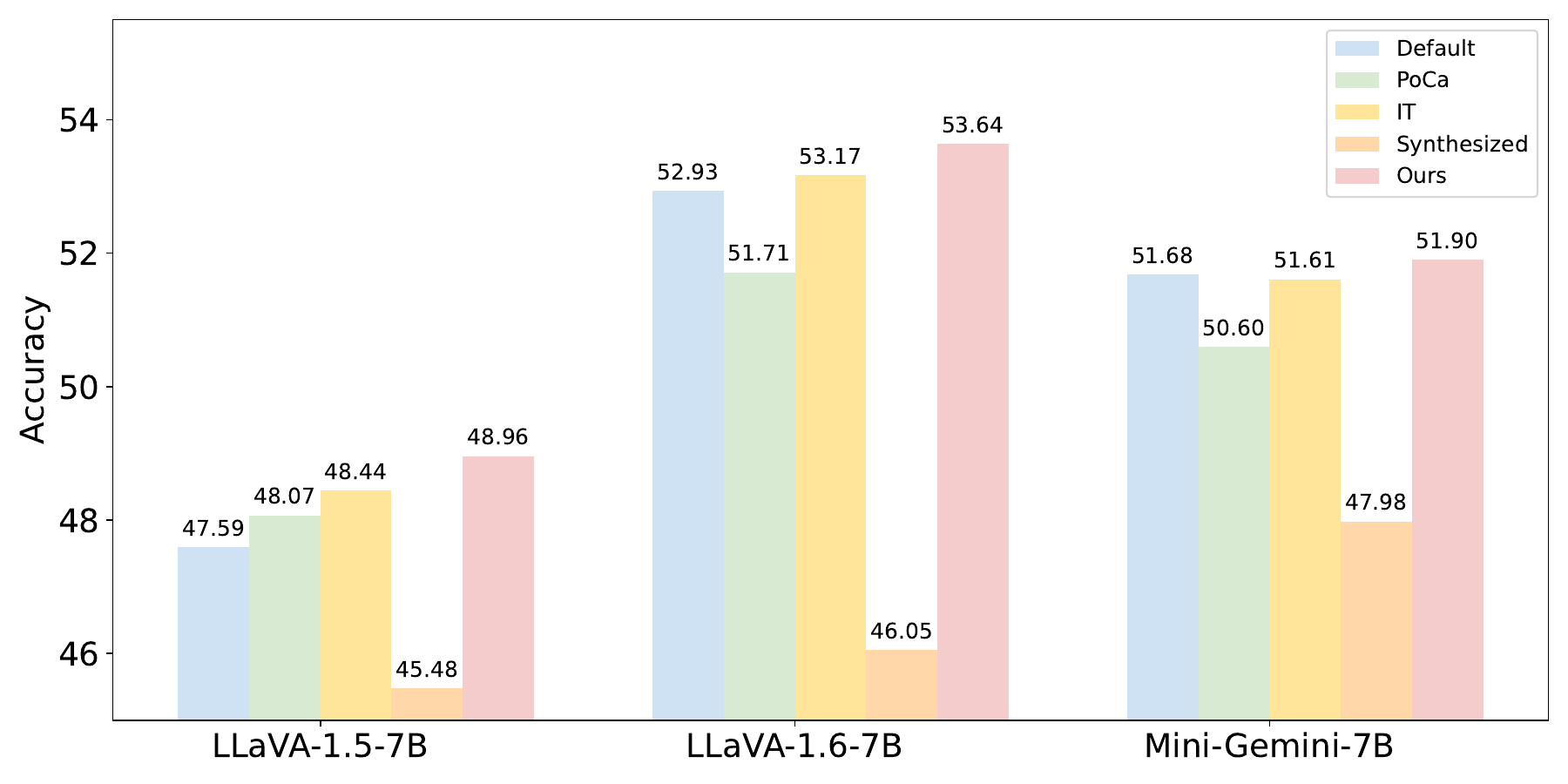}
    
   \caption{The performance of different methods and models on the VQA task shows that our method achieves the best results.}
   \label{fig:vqa}
\end{figure}

\begin{table*}[h]
    \centering
    \renewcommand{\arraystretch}{0.86}
     \setlength{\tabcolsep}{3.95pt}
    \begin{tabular}{c|cccc|cccc|cccc}
        \toprule
        \multirow{2}{*}{Tuning Data}  &  \multicolumn{4}{c}{Adversarial} & \multicolumn{4}{c}{Random} & \multicolumn{4}{c}{Popular} \\
        &Acc& Precision &Recall& F1 &Acc& Precision &Recall& F1 &Acc& Precision &Recall& F1  \\
        \midrule
        /  & 82.70 &	85.62 	&78.60 	&81.96 &87.67 	&95.35 	&79.20 &	86.53 &85.73 &	91.10 &	79.20 &	84.74  \\
        \{LLaVA\} & 83.77 &	88.00 &	78.20 &	82.81 &87.83 &	96.86 &	78.20 &	86.54 &87.07 &	\textbf{95.06} &	78.20& 	85.81  \\
        Ours-\{LLaVA\}  & \textbf{84.37} &	\textbf{88.10} &	\textbf{79.47 }&	\textbf{83.56} &\textbf{88.60} &	\textbf{97.23} &	\textbf{79.47} 	&\textbf{87.45}  &\textbf{87.33}	&94.30 	&\textbf{79.47}	&\textbf{86.25} \\
        \bottomrule
    \end{tabular}
    \caption{LLaVA-1.5-7B performance with and without fine-tuning on synthesized detailed caption data on the POPE benchmark. ``\{LLaVA\}" refers to detailed captions generated directly by LLaVA-1.5-7B, while ``Ours-\{LLaVA\}" refers to the data constructed using our method. ``Acc" denotes accuracy, and ``F1" denotes the F1 score.}
    \label{tab:pope_results}
\end{table*}
\begin{table*}[h]
    \centering
    \renewcommand{\arraystretch}{0.86}
     \setlength{\tabcolsep}{3.7pt}
    \begin{tabular}{l|c|cccc|ccccc}
        \toprule
        GroundTruth&Tuning Data & BLEU-1 & BLEU-2 & BLEU-3 & BLEU-4 & CIDEr & METEOR & ROUGE & SPICE & WMD \\  
        \midrule
         \multirow{3}{*}{GT-\{LLaVA\}} & / &9.16  & 5.70 & 3.37 & 2.07 & 0.47 & 10.62 & 19.54 & 17.65& 41.37  \\
        & \{LLaVA\} & 10.02 &	6.25 &	3.79 &	2.40 &	0.00& 	11.05 	&20.19& 	18.26 &	41.81 \\
        & Ours-\{LLaVA\} & \textbf{34.09} 	&\textbf{19.33} &	\textbf{10.91} &	\textbf{6.47} &	\textbf{5.63} &	\textbf{17.34} &	\textbf{23.35 }&\textbf{	20.35} 	&\textbf{43.11} \\
         \midrule
         \multirow{3}{*}{GT-\{GPT4-V\}} & / &  7.50 &	4.05 &	1.94 &	0.98 	&0.00 &	9.11 &	15.67 &	13.29 &	37.74 \\
         & \{LLaVA\} & 9.05 	&4.78 &	2.32 &	1.20 &	0.00 &	9.71 &	16.31 &	14.12 	&37.83 \\
        & Ours-\{LLaVA\} &\textbf{27.68} &	\textbf{13.84} &	\textbf{6.68} &	\textbf{3.50} 	&\textbf{3.11} &	\textbf{15.13} 	&\textbf{19.75 }	&\textbf{16.30} 	&\textbf{39.26} \\
        \bottomrule
    \end{tabular}
    \caption{LLaVA-1.5-7B performance with and without fine-tuning on synthesized detailed caption data on the DID-Bench, with GT-\{LLaVA\} and GT-\{GPT4-V\} as ground truth captions generated by LLaVA and GPT-4Vision, respectively. ``\{LLaVA\}" refers to detailed captions generated directly by LLaVA-1.5-7B, while ``Ours-\{LLaVA\}" refers to the data constructed using our method.}
    \label{tab:did_results}
\end{table*}

\subsection{VQA Task}
As shown in \cref{fig:vqa}, our method consistently enhances accuracy across models, with improvements from 0.22\% to 1.37\%, reflecting higher informational richness and enabling more comprehensive responses. In contrast, some methods introduce errors or omit critical information, leading to incorrect responses.

\subsection{Fintune Result in POPE and DID-Bench}

\noindent \textbf{Experiment Settings.} We fine-tuned the LLaVA-1.5-7B model using LoRA \cite{lora} with the default pipeline parameters. The learning rate was set to $2e^{-4}$, the LoRA rank was 128, and the scaling factor was 256. For fine-tuning, we used 10k image-text pairs sourced from COCO \cite{lin2014COCO-data}, VG \cite{krishna2017visual}, and SAM \cite{kirillov2023SAM}, which were annotated using our method based on LLaVA-1.5-7B. We then compared the performance of our annotations with the directly annotated data from LLaVA.

\noindent \textbf{POPE and DID-Bench Results.} We evaluated the impact of our annotation method on model performance by conducting experiments on two benchmarks: POPE \cite{li2023pope} and DID-Bench \cite{pi2024image-textual}. As shown in the results of POPE \cref{tab:pope_results}, even with only 10k annotated pairs for fine-tuning, our approach significantly mitigates hallucinations compared to the baseline model, which uses direct annotations. While direct annotations provide some improvement, our method, which uses higher-quality annotations, yields more substantial gains. In \cref{tab:did_results}, we show the results on the DID-Bench benchmark, where fine-tuning with captions generated by our method substantially improves the model's ability to generate high-quality captions, outperforming the baseline.

 % \begin{tabular}{l|c|cccc|ccccc} % 将 tabular* 替换为 tabular
 %        \toprule
 %        GroundTruth & Method & BLEU-1 & BLEU-2 & BLEU-3 & BLEU-4 & CIDEr & METEOR & ROUGE & SPICE & WMD \\
 %        \midrule
 %        \rowcolor{gray!20}
 %        \multicolumn{11}{c}{\textit{\textbf{LLaVA-1.5-7B}}} \\
 %        \multirow{5}{*}{GT-\{LLaVA\}} & Global & 9.16  & 5.70 & 3.37 & 2.07 & 0.47 & 10.62 & 19.54 & 17.65& 41.37 \\

\section{More Case Studies}

\begin{figure*}
  \centering
   \includegraphics[width=0.95\linewidth]{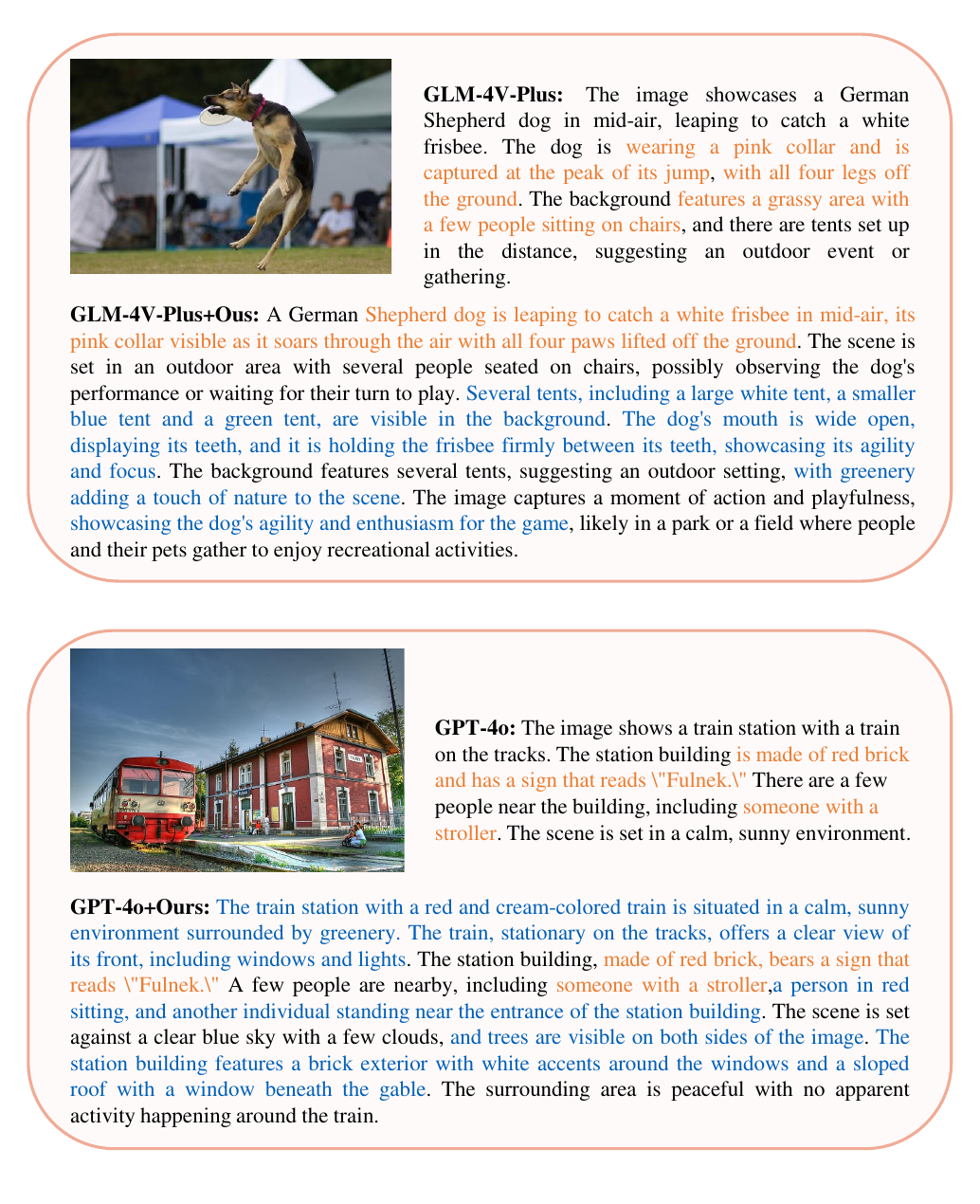}
    \caption{Visualization of the original description and the modified description. \textcolor{orange}{Shared information} and \textcolor{cyan}{newly added details} are highlighted in different colors for clarity.}
    \label{fig:case1}
\end{figure*}

\begin{figure*}
  \centering
   \includegraphics[width=0.95\linewidth]{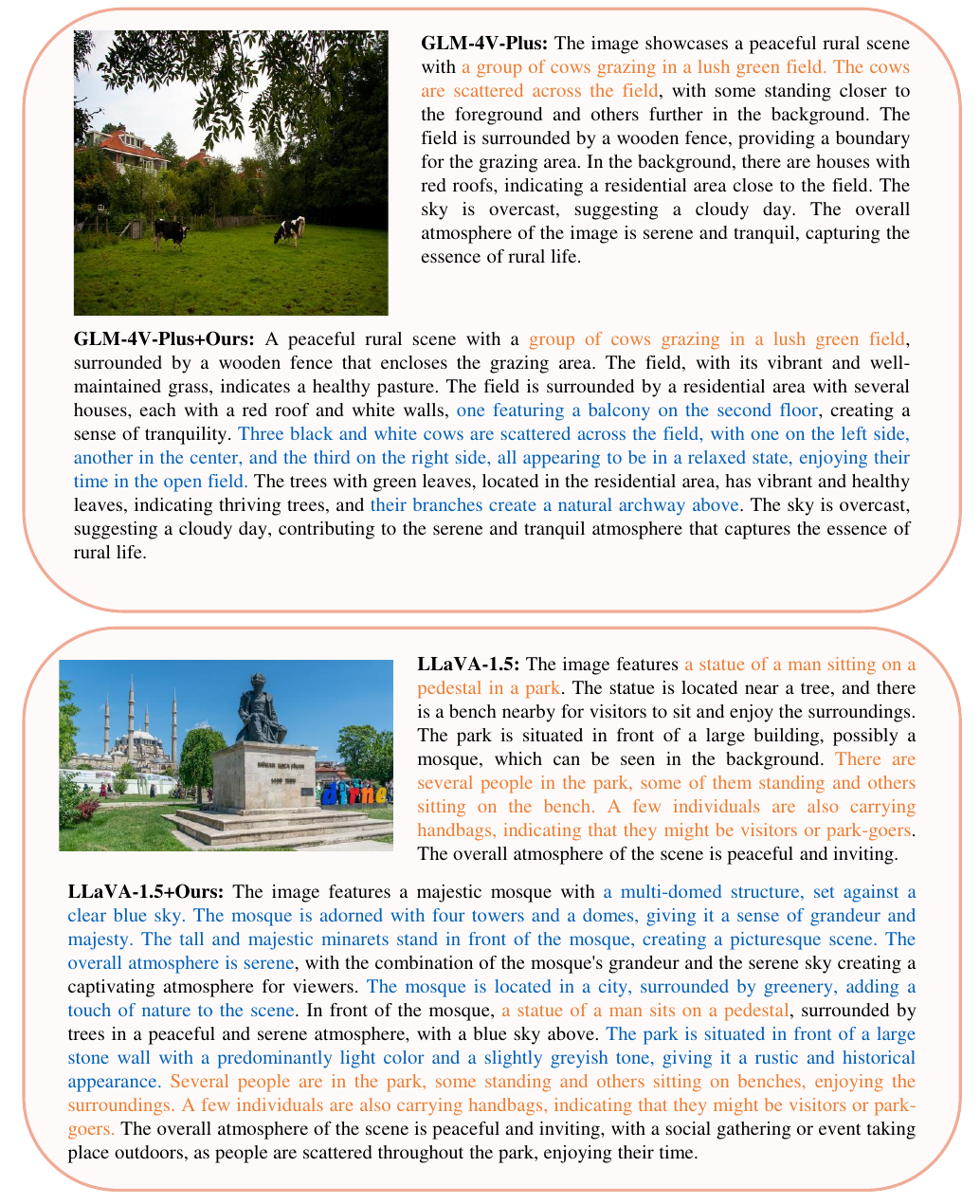}
    \caption{Visualization of the original description and the modified description. \textcolor{orange}{Shared information} and \textcolor{cyan}{newly added details} are highlighted in different colors for clarity.}
    \label{fig:case2}
\end{figure*}

We provide more qualitative comparisions between MLLM-generated and our-generated image descriptions in \cref{fig:case1} and \cref{fig:case2}.
\newpage



% \begin{figure*}
%   \centering
%    \includegraphics[width=1\linewidth]{sec/case3.pdf}
%     \caption{Visualization of the original description and the modified description. \textcolor{orange}{Shared information} and \textcolor{cyan}{newly added details} are highlighted in different colors for clarity.}
%     \label{fig:case2}
% \end{figure*}

\newpage